\documentclass[10pt,journal,compsoc]{IEEEtran}
\usepackage{ifpdf}
\ifpdf
\else
\fi
\ifCLASSINFOpdf
   \usepackage[pdftex]{graphicx}
   \graphicspath{{../pdf/}{../jpeg/}}
   \DeclareGraphicsExtensions{.pdf,.jpeg,.png}
\else
   \usepackage[dvips]{graphicx}
   \graphicspath{{../eps/}}
   \DeclareGraphicsExtensions{.eps}
\fi
\usepackage{amsmath}
\usepackage{array}

\ifCLASSOPTIONcompsoc
 \usepackage[caption=false,font=footnotesize,labelfont=sf,textfont=sf]{subfig}
\else
 \usepackage[caption=false,font=footnotesize]{subfig}
\fi

\usepackage{stfloats}
\ifCLASSOPTIONcaptionsoff
 \usepackage[nomarkers]{endfloat}
\let\MYoriglatexcaption\caption
\renewcommand{\caption}[2][\relax]{\MYoriglatexcaption[#2]{#2}}
\fi
\let\MYorigsubfloat\subfloat
\renewcommand{\subfloat}[2][\relax]{\MYorigsubfloat[]{#2}}
\usepackage{xspace}
\makeatletter
\DeclareRobustCommand\onedot{\futurelet\@let@token\@onedot}
\def\@onedot{\ifx\@let@token.\else.\null\fi\xspace}

\def\eg{\emph{e.g}\onedot} 
\def\ie{\emph{i.e}\onedot}

\makeatother

\usepackage{enumitem}
\usepackage{caption}
\usepackage{times}
\usepackage{epsfig}
\usepackage{graphicx}
\usepackage{amsmath}
\usepackage{amssymb}
\usepackage{eso-pic}
\usepackage{helvet}
\usepackage{courier}
\usepackage{url}
\usepackage{mathrsfs}
\usepackage{multirow}
\usepackage{comment}
\usepackage{marvosym}
\usepackage{bbm}
\usepackage{enumerate}
\usepackage{soul}
\usepackage[utf8]{inputenc}
\usepackage{amsfonts}
\usepackage{nicefrac}
\usepackage{microtype}
\usepackage{booktabs} 
\usepackage{algorithm}
\usepackage{algorithmicx}
\usepackage{algpseudocode}
\usepackage{makecell}
\usepackage{multirow}

\usepackage{bbding}
\usepackage[numbers]{natbib}
\usepackage[breaklinks=true,bookmarks=false,colorlinks]{hyperref}
\usepackage{cleveref}

\usepackage[normalem]{ulem}

\newcommand{\eqn}[1]{Eq.~\eqref{#1}}
\newcommand{\sect}[1]{Section~\ref{#1}}
\newcommand{\tab}[1]{Table~\ref{#1}}
\newcommand{\fig}[1]{Fig.~\ref{#1}}
\newcommand{\alg}[1]{Alg.~\ref{#1}}

\newcommand{\myparagraph}[1]{\vspace{5pt} \noindent \textbf{#1}}

\definecolor{MyDarkBlue}{rgb}{0,0.5,1}
\definecolor{MyDarkGreen}{rgb}{0.02,0.6,0.02}
\definecolor{MyDarkRed}{rgb}{0.8,0.02,0.02}
\definecolor{MyDarkOrange}{rgb}{0.40,0.2,0.02}
\definecolor{MyPurple}{RGB}{111,0,255}
\definecolor{MyRed}{rgb}{1.0,0.0,0.0}
\definecolor{MyGold}{rgb}{0.75,0.6,0.12}
\definecolor{MyDarkgray}{rgb}{0.66, 0.66, 0.66}
\newcommand{\model}{Iso-Dream++}

\algdef{SE}[DOWHILE]{Do}{doWhile}{\algorithmicdo}[1]{\algorithmicwhile\ #1}%

\begin{document}

\title{Model-Based Reinforcement Learning with Isolated Imaginations}

%
%
%
%

\author{Minting~Pan,
        Xiangming~Zhu,
        Yitao~Zheng,
        Yunbo~Wang,
        Xiaokang~Yang,~\IEEEmembership{Fellow,~IEEE}
\IEEEcompsocitemizethanks{
\IEEEcompsocthanksitem The authors are with MoE Key Lab of Artificial Intelligence, AI Institute, Shanghai Jiao Tong University, China.
\IEEEcompsocthanksitem Corresponding author: Y. Wang, yunbow@sjtu.edu.cn.
\IEEEcompsocthanksitem Code: \url{https://github.com/panmt/MBRL_with_Isolated_Imaginations}.
}
}

\markboth{IEEE Transactions on Pattern Analysis and Machine Intelligence,~Vol.~XX, No.~X, March~2023}%
{Pan \MakeLowercase{\textit{et al.}}: Iso-Dream++: Model-Based Reinforcement Learning with Isolated Imaginations}

\IEEEtitleabstractindextext{%

\begin{abstract}

World models learn the consequences of actions in vision-based interactive systems. However, in practical scenarios like autonomous driving, noncontrollable dynamics that are independent or sparsely dependent on action signals often exist, making it challenging to learn effective world models. To address this issue, we propose Iso-Dream++, a model-based reinforcement learning approach that has two main contributions. First, we optimize the inverse dynamics to encourage the world model to isolate controllable state transitions from the mixed spatiotemporal variations of the environment. Second, we perform policy optimization based on the decoupled latent imaginations, where we roll out noncontrollable states into the future and adaptively associate them with the current controllable state. This enables long-horizon visuomotor control tasks to benefit from isolating mixed dynamics sources in the wild, such as self-driving cars that can anticipate the movement of other vehicles, thereby avoiding potential risks. On top of our previous work~\cite{paniso}, we further consider the sparse dependencies between controllable and noncontrollable states, address the training collapse problem of state decoupling, and validate our approach in transfer learning setups. Our empirical study demonstrates that Iso-Dream++ outperforms existing reinforcement learning models significantly on CARLA and DeepMind Control.

\end{abstract}

}

\maketitle

\IEEEdisplaynontitleabstractindextext

%
\IEEEpeerreviewmaketitle

\section{Introduction}

\begin{figure*}[t]
\centerline{\includegraphics[width=0.95\linewidth]{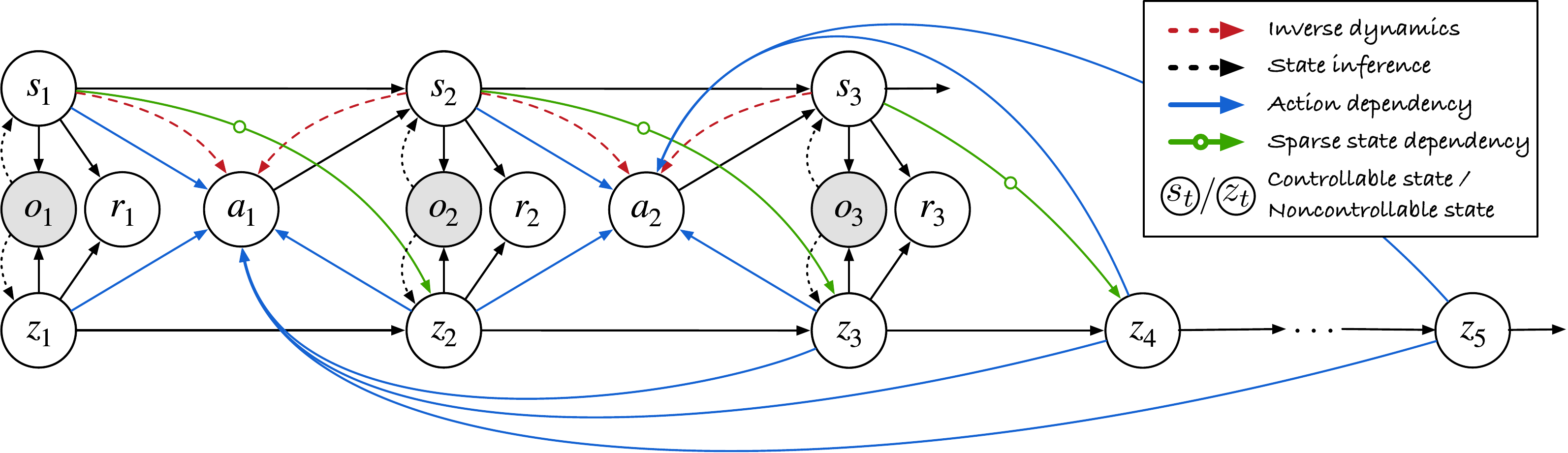}}
\caption{Graphic model of our approach. The world model learns to decouple mixed visual dynamics into controllable states ($s_t$) and noncontrollable states ($z_t$) by optimizing the inverse dynamics (\textcolor{MyDarkRed}{as indicated by the red dashed arrows}). With state decoupling, the RL agent can make decisions based on the forecasts of future noncontrollable dynamics of the environment (\textcolor{MyDarkBlue}{blue arrows}).
In forward modeling, we consider the sparse dependency of next-step noncontrollable states on current controllable states (\textcolor{MyDarkGreen}{green arrows}). 
In representation learning, we further cope with the imbalance of dynamic information learned in different state transition branches.
}
\label{fig:intro}
\vskip -1mm
\end{figure*}

Humans can infer and predict real-world dynamics by simply observing and interacting with the environment. 
Inspired by this, many cutting-edge AI agents use self-supervised learning \cite{oh2015action,ha2018world,ebert2018visual} or reinforcement learning \cite{oh2017value,hafner2019dream,sekar2020planning} techniques to acquire knowledge from their surroundings. 
Among them, world models \cite{ha2018world} have received widespread attention in the field of robotic visuomotor control, and led the recent progress in model-based reinforcement learning (MBRL) with visual inputs \cite{hafner2019dream,sekar2020planning,hafner2020mastering,kaiser2020model}.
One representative approach called Dreamer~\cite{hafner2019dream} learns a differentiable simulator of the environment (\ie, the world model) using observations and actions of an actor-critic agent, then updates the agent by optimizing its behaviors based on future latent states and rewards (\ie, latent imagination) generated by the world model.
However, since the observation trajectories are high-dimensional, highly non-stationary, and often driven by multiple sources of physical dynamics, how to learn effective world models in complex visual scenes remains an open problem.

In this paper, we propose to understand the world by decomposing it into \textit{controllable} and \textit{noncontrollable} state transitions, \ie, $s_{t+1}\sim p(\cdot \mid s_t,a_t)$ and $z_{t+1}\sim p(\cdot \mid z_t)$, according to the responses to action signals.
This idea is largely inspired by practical scenarios such as autonomous driving, in which we can naturally divide spatiotemporal dynamics in the system into controllable parts that perfectly respond to the actions (\eg, accelerating and steering) and parts beyond the control of the agent (\eg, movement of other vehicles).
Decoupling latent state transitions in this way can improve MBRL in three aspects: 
\begin{itemize}[leftmargin=*]
  \item It allows decisions to be made based on predictions of future noncontrollable dynamics that are independent (or indirectly dependent) of the action, thereby improving the performance on long-horizon control tasks. For example, in the CARLA self-driving environment, potential risks can be better avoided by anticipating the movement of other vehicles.
  \item Modular world models improve the robustness of the RL agent in noisy environments, as demonstrated in our modified DeepMind Control Suite with the time-varying background.
  \item The isolation of controllable state transitions further facilitates transfer learning across different but related domains. We can adapt parts of the world model to novel domains based on our prior knowledge of the domain gap. 
\end{itemize}

Specifically, we present \model{}, a novel MBRL framework that learns to decouple and leverage the controllable and noncontrollable state transitions.
Accordingly, it improves the original Dreamer \cite{hafner2019dream} from two perspectives: (i) \textit{a new form of world model representation} and (ii) \textit{a new actor-critic algorithm to derive the behavior from the world model.}

\subsection{How to learn a decoupled world model?}

From the perspective of representation learning, we improve the world model to separate mixed visual dynamics into an action-conditioned branch and an action-free branch of latent state transitions (see \fig{fig:intro}). 
These components are jointly trained to maximize the variational lower bounds.
Besides, the action-conditioned branch is particularly optimized with \textit{inverse dynamics} as an additional objective function, that is, to reason about the actions that have driven the ``controllable'' state transitions between adjacent time steps.

Nonetheless, as we have observed in our preliminary work at NeurIPS'2022~\cite{paniso}, which we call Iso-Dream, the learning process of inverse dynamics is prone to the problem of ``training collapse'', where the action-conditioned branch captures all dynamic information, while the action-free branch learns almost nothing. 
To further isolate different dynamics in an unsupervised manner, we use new forms of min-max variance constraints to regularize the information flow of dynamics in the decoupled world model.
More concretely, we provide a batch of hypothetical actions to the world model, and encourage the action-conditioned branch to produce different state transitions based on the same state, while penalizing the diversity of those in the action-free branch.

\subsection{How to improve behavior learning based on decoupled world models?}

Humans can decide how to interact with the environment at each moment based on their anticipation of future changes in their surroundings. 
Accordingly, by decoupling the state transitions, our approach can explicitly forecast the evolution of action-independent dynamics in the system, thereby greatly benefiting downstream decision-making tasks.
Unlike Dreamer, it performs latent state imagination in both the training phase and testing phase of the agent behaviors to make more forward-looking decisions.
As shown by the blue arrows in \fig{fig:intro}, the policy network integrates the current controllable state and multiple steps of predicted noncontrollable states through an attention mechanism. 
Intuitively, since future noncontrollable states at different steps may have different weights of impact on the current decision of the agent, the attention mechanism enables the agent to adaptively consider possible future interactions with the environment. It ensures that only appropriate future states are fed back into the policy.

Despite the effectiveness of the new behavior learning scheme, it only considers the indirect influence of action-free dynamics on future action-conditioned dynamics through agent behaviors (\ie, $z_{t:t+\tau} \rightarrow a_t \rightarrow s_{t+1}$). 
Another improvement of our approach over Iso-Dream is that it further models the \textit{sparse dependency} of future noncontrollable states on current controllable states (\ie, $s_{t} \rightarrow z_{t+1}$), which is indicated by the green arrows in \fig{fig:intro}. 
In practical scenarios, for example, when we program a robot to compete with another one in a dynamic game, the opponent can adjust its policy according to the behavior of our agent.
In autonomous driving, when an agent vehicle veers into the lane of other vehicles, typically those vehicles will slow down to avoid a collision.
Because of the proposed solution to training collapse, modeling the sparse dependency does not affect the disentanglement learning ability of the world model.
In behavior learning, actions are sampled from $\pi(s_t,z_{t:t+\tau})$, where $z_{t+1}\sim p(\cdot|z_t,s_t)$, while due to the sparsity of the cross-branch dependencies, the long-horizon noncontrollable future can be approximated as $z_{t+2:t+\tau}\sim p(\cdot|z_{t+1:t+\tau-1})$.

We evaluate \model{} in the following domains: (i) the CARLA autonomous driving environment in which other vehicles can be naturally viewed as noncontrollable components; (ii) the modified DeepMind Control Suite with noisy video background. 
Our approach outperforms existing approaches by large margins and further achieves significant advantages in transfer learning by isolating dynamics.
It can selectively transfer controllable or noncontrollable parts of the learned state transition functions from the source domain to the target domain according to the prior information.

The main contributions of this paper are summarized as follows:
\begin{itemize}[leftmargin=*]
  \item We present a new world model and encourage the decomposition of latent state transitions by optimizing \textit{inverse dynamics}. 
  \item We introduce the \textit{min-max variance constraints} to prevent all information from collapsing into a single state transition branch.
  \item We improve the actor-critic algorithm to make \textit{forward-looking decisions} based on the forecasts of future noncontrollable states.
  \item We model the \textit{sparse dependency} of the next-step noncontrollable dynamics on current controllable dynamics to provide a more accurate simulation of some practical dynamic environments.
  \item We empirically demonstrate the advantages of \model{} over existing methods in standard, noisy, and \textit{transfer learning} setups.
\end{itemize}
In summary, we extend our previous studies with (i) the min-max variance constraints, (ii) the sparse dependence between the decoupled latent states, and (iii) the transfer learning experiments.

\section{Problem Overview}

\begin{figure*}[t] 
    \centering
	  \subfloat[]{
       \includegraphics[width=3.2in]{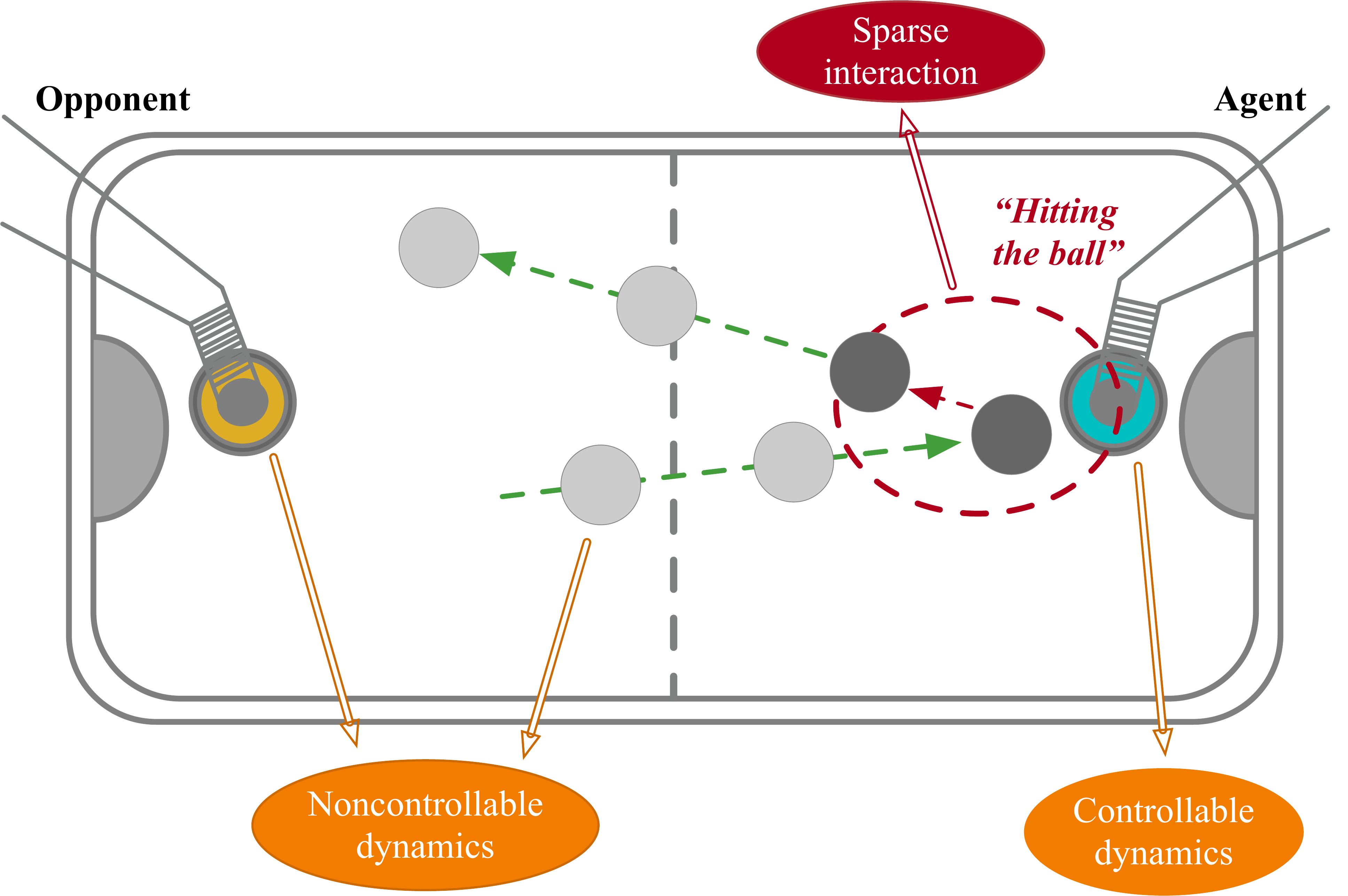}}
    \label{fig:ball_sparse_vis} 
    \hfil
	  \subfloat[]{
        \includegraphics[width=2.9in]{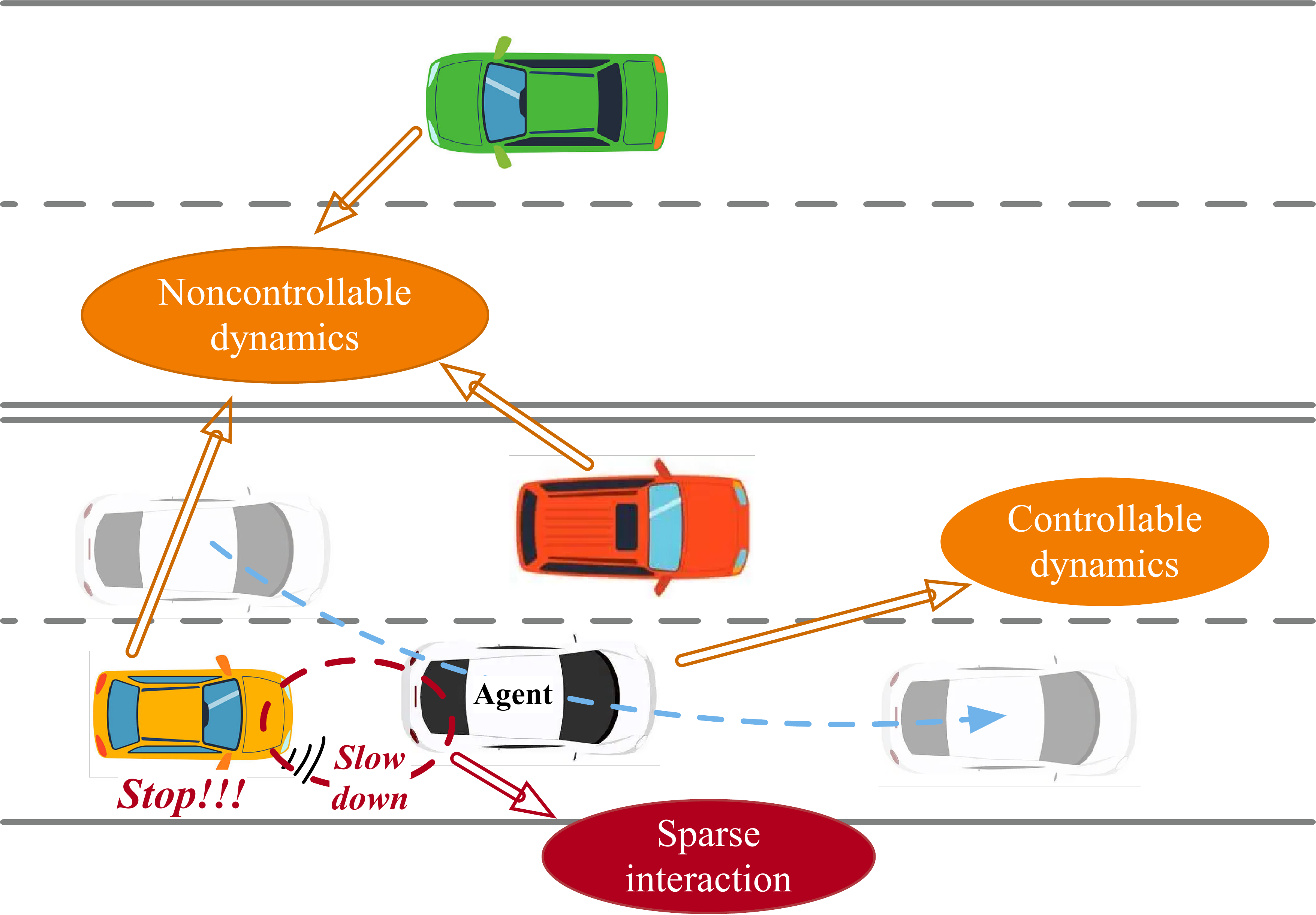}}
    \label{fig:car_sparse_vis}
    \vspace{-5pt}
	\caption{Examples of sparse dependency of the noncontrollable state on the controllable state. (a) A game resembling ice hockey that is played on a desk, in which the ego-agent is controllable, while the opponent robot can be seen as the noncontrollable part, and the hockey puck can be mostly considered to have predictable dynamics independent of the agent's actions. \textit{Sparse dependency} occurs at the moment the agent hits the puck because it may change direction depending on the agent's current state. (b) In autonomous driving, other vehicles (\textit{yellow}) will change their driving directions to avoid collision when the ego-agent (\textit{white}) takes up their driveway.}
	  \label{fig:sparse_vis} 
   \vspace{-5pt}
\end{figure*}

\subsection{Problem Definition}

In visual control tasks, the agent learns the action policy directly from high-dimensional observations. We formulate visual control as a partially observable Markov decision process (POMDP) with a tuple $(\mathcal S,\mathcal A,\mathcal T,\mathcal R,\mathcal O)$, where $\mathcal S$ is the state space, $\mathcal A$ is the action space, $\mathcal O$ is the observation space, $\mathcal R(s_t, a_t)$ is the reward function, and $ \mathcal T(s_{t+1} \mid s_t, a_t) $ is the state-transition distribution. 
At each timestep $t\in[1; T]$, the agent takes an action $a_t \in A$ to interact with the environment and receives a reward $r_t = \mathcal R(s_t, a_t)$. The objective is to learn a policy that maximizes the expected cumulative reward $\mathbb E_p[\sum^T_{\tau=1} r_{\tau}]$.
In this setting, the agent cannot access the true states in $\mathcal S$.

\subsection{Key Challenges}
\label{sec:challenges}

\myparagraph{Challenge 1: How to learn future-conditioned policies without the expensive Monte-Carlo planning?}
Forecasting future environmental changes is useful for decision-making in a non-stationary system.
A typical solution, such as the cross-entropy method (CEM), is to perform Monte-Carlo sampling over future actions and value the consequences of multiple action trajectories \cite{finn2017deep,ebert2018visual,hafner2019learning}.
These algorithms are expensive in computational cost, especially when we have large action and state spaces.
The question is: \textit{Can we design an RL algorithm that allows for future-conditioned decision-making without playing dice in the action space?}

\myparagraph{Challenge 2: How to avoid ``training collapse'' in unsupervised dynamics disentanglement?} 
Despite the great success in unsupervised representation learning \cite{locatello2019challenging,he2020momentum,qian2022unsupervised}, it remains a challenge to disentangle the controllable and noncontrollable dynamic patterns in non-stationary visual scenes. 
One potential solution is to employ modular structures that learn different dynamics in separate branches. However, without proper constraints, the model may suffer from ``training collapse'', where one branch captures all useful information and the other learns almost nothing.
This phenomenon may occur when the noncontrollable dynamics components are easy to predict. In this case, we consider adding further constraints to the learning objects of the action-conditioned and action-free state transition branches, encouraging them to isolate the noncontrollable part from the mixed dynamics.

\myparagraph{Challenge 3: How to model situations where the agent behavior has only a sparse/indirect impact on noncontrollable dynamics?}
As we know, in realistic scenarios, the noncontrollable component of the dynamics may not evolve independently but may depend on the motions of the controllable component. 
For instance, in \fig{fig:sparse_vis} (a), the hockey puck on the desk (noncontrollable part) changes its direction when the agent (controllable part) hits it. 
For autonomous driving, in \fig{fig:sparse_vis} (b), other vehicles (noncontrollable part) will slow down to avoid a collision when the agent (controllable part) takes their lane. 
If we assume that our actions do not indirectly affect other vehicles on the road, then for safety reasons, a sub-optimal policy for handling heavy traffic could be to follow the vehicle in front of us instead of changing lanes.
Accordingly, we propose a sparse dependency mechanism that enhances our model's decision-making ability. 
Empirical results are illustrated in \fig{fig:sparse_ablation_vis}.

\subsection{Basic Assumptions}
\label{sec:motivation}

In our proposed framework as shown in \fig{fig:intro}, when the agent receives a sequence of visual observations $o_{1:T}$, the underlying spatiotemporal dynamics can be defined as $u_{1:T}$. The evolution of different dynamics can be caused by different forces, but here we aim to decouple $u_{1:T}$ into controllable latent states $s_{1:T}$ and time-varying noncontrollable latent states $z_{1:T}$, such that:
\begin{equation}
\label{eq:basic_assum}
  \begin{aligned}
    u_{1:T} &\sim  (s, z)_{1:T}, \\
    s_{t+1} &\sim p(s_{t+1} \mid s_t, a_t), \\
    z_{t+1} &\sim p(z_{t+1} \mid z_t),
  \end{aligned}
\end{equation}
where $a_t$ is the action signal. 
By isolating $s_t$ and $z_t$ to each other, we model their state transitions of $p(s_{t+1} \mid s_t, a_t)$ and $p(z_{t+1} \mid z_t)$ respectively. 
We assume that a more clear decoupling of $s_t$ and $z_t$ can benefit both long-term predictions and decision-making.
As an extension of our preliminary work \cite{paniso}, we additionally model the sparse dependency of noncontrollable dynamics on controllable dynamics (as described below). Thus, when a sparse event is detected, the transition of noncontrollable state in \eqn{eq:basic_assum} can be rewritten as $z_{t+1} \sim p(z_{t+1} \mid z_t, s_t)$.

It assumes that the agent can greatly benefit from predicting the consequences of external noncontrollable forces.
During behavior learning, we roll out the noncontrollable states and then associate them with the current controllable states for more proactive decision-making.
We derive the action policy by
\begin{equation}  
    a_t \sim \pi(a_t \mid s_t, \mathbbm{1} \odot z_{t:t+\tau}),
\end{equation}
where $\mathbbm{1}$ is an indicator according to our prior knowledge about the environment. For example, in autonomous driving, since it is reasonable for the ego-agent to make decisions based on the predictions about the future states of other vehicles, we have $\mathbbm{1}=1$ and calculate the relations between $s_t$ and the imagined noncontrollable states in a time horizon $\tau$. 
Otherwise, for some specific tasks where the noncontrollable components are irrelevant to decision-making, we can simply set the indicator function to $\mathbbm{1}=0$ and treat them as noisy distractions.

\section{Method}

\begin{figure*}[t]
    \centering
    \includegraphics[width=0.9\linewidth]{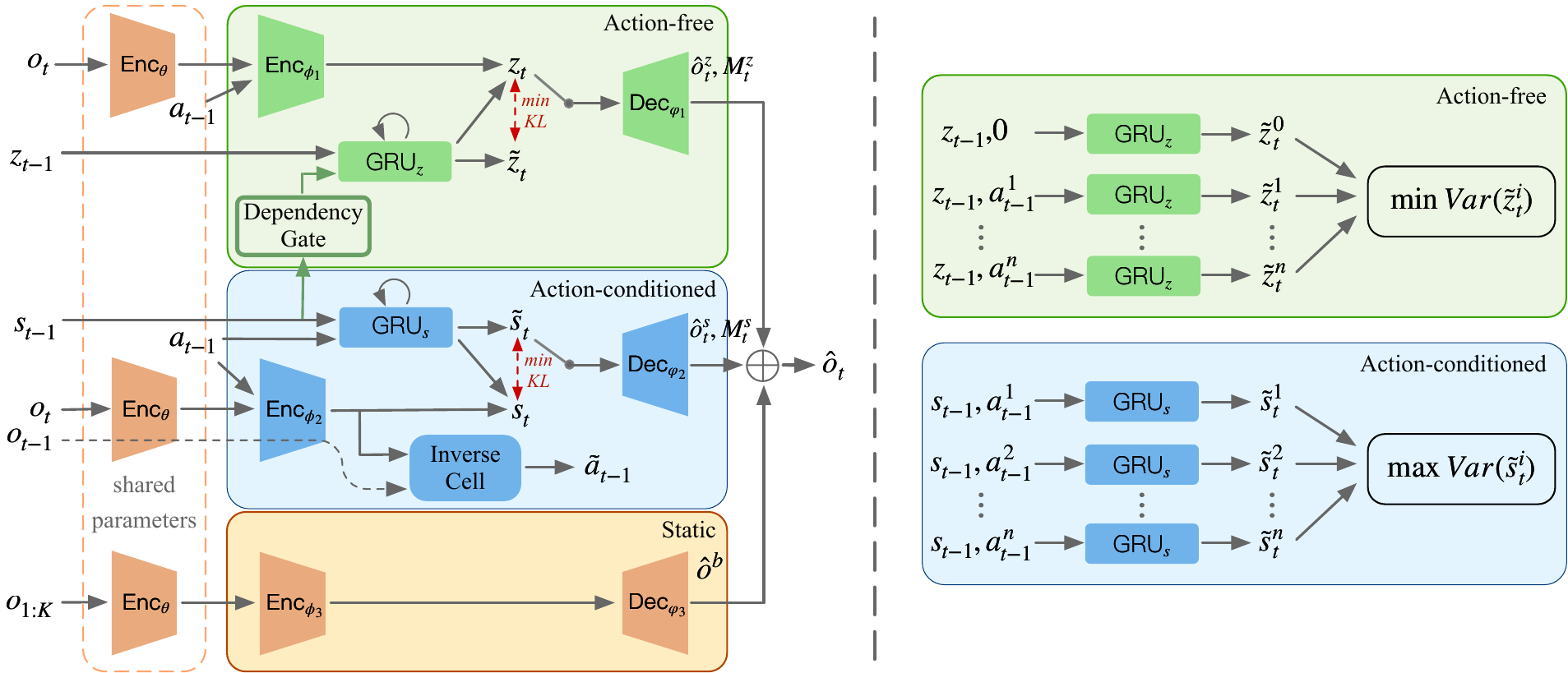}
    \caption{The overall architecture of the world model in \model{}. \textbf{Left:} The world model has three branches to explicitly disentangle controllable and noncontrollable state transitions, as well as the static components from visual data. 
    \textbf{Right:} Illustration of calculating variance in different branches. Given the different action signals, our objective is to minimize the diversity of state transitions in the action-free branch and maximize the diversity of those in the action-conditioned branch. 
   }
    \label{fig:rl_model_and_policy}
    \vspace{-5pt}
\end{figure*}

In this section, we present the technical details of \model{} for decoupling and leveraging controllable and noncontrollable dynamics for visual MBRL. 
The overall pipeline is based on Dreamer \cite{hafner2019dream}, where we learn the world model from a dataset of past experience, learn behaviors from imagined sequences of compact model states, and execute the behavior policy in the environment to grow the experience dataset.

In \sect{sec:inverse_dynamics}, we first introduce the three-branch world model and its training objectives of \textit{inverse dynamics}. 
In \sect{sec:variance}, we propose the \textit{min-max variance constraints} to regularize the dynamics representations in each state transition branch to enhance disentanglement learning and avoid training collapse. 
In \sect{sec:spare-dependency}, we present a network structure to model the phenomenon that future noncontrollable dynamics can be sparsely influenced by current controllable dynamics.
In \sect{sec:behavior-learning}, we present an actor-critic method that is trained on the imaginations of the decoupled world model latent states, so that the agent may consider possible future states of noncontrollable dynamics in behavior learning.
Finally, in \sect{sec:Policy-Deployment}, we discuss how our model is deployed to interact with the environment.

\subsection{World Models with Dynamics Isolation}
\label{sec:wm}

Inspired by prior research \citep{slotattention,rim} that demonstrates the efficacy of modular structures for disentanglement learning, we use an architecture with multiple branches to model different dynamics independently, according to their respective physical laws. 
Each individual branch tends to present robust features, even when the dynamic patterns in other branches undergo changes.
Specifically, our three-branch model, illustrated in the left panel of \fig{fig:rl_model_and_policy}, disentangles visual observations into controllable dynamics state $s_t$, noncontrollable dynamics state $z_t$, and a time-invariant component of the environment.
The action-conditioned branch models the controllable state transition $p(s_{t+1} \mid s_t, a_t)$. 
It follows the RSSM architecture from PlaNet \citep{hafner2019learning} to use a recurrent neural network $\texttt{GRU}_s(\cdot)$, the deterministic hidden state $h_t$, and the stochastic state $s_t$ to form the transition model, where the GRU keeps the historical information of the controllable dynamics. 
The action-free branch models $p(z_{t+1} \mid z_t)$ with similar network structures. The transition models with separate parameters can be written as follows:
\begin{equation}
  \begin{split}
  p(\tilde{s}_t \mid s_{<t}, a_{<t}) = p(\tilde{s}_t \mid h_t), \\ 
  p(\tilde{z}_t \mid z_{<t}) = p(\tilde{z}_t \mid h_t^\prime), \\ 
  \end{split}
\end{equation}
where $ h_t = \texttt{GRU}_s(h_{t-1}, s_{t-1}, a_{t-1}), \ h_t^\prime = \texttt{GRU}_z(h_{t-1}^\prime, z_{t-1})$.
We here use $\tilde{s}_t$ and $\tilde{z}_t$ to denote the prior representations.
We optimize the transition models with posterior representations that are derived from $s_t \sim q(s_t \mid h_{t}, o_t, a_{t-1})$ and $z_t \sim q(z_t \mid h_{t}^\prime, o_t, a_{t-1})$. 
We learn the posteriors from the observation at current time step $o_t \in \mathbb{R}^{3 \times H \times W}$ by a shared encoder $\texttt{Enc}_{\theta}$ and subsequent branch-specific encoders $\texttt{Enc}_{\phi_1}$ and $\texttt{Enc}_{\phi_2}$.
Notably, we feed actions into both $\texttt{Enc}_{\phi_1}$ and $\texttt{Enc}_{\phi_2}$, which differs from our previous work \citep{paniso}. In the static branch, where there is no state transition, we only use an encoder $\texttt{Enc}_{\phi_3}$ and a decoder $\texttt{Dec}_{\varphi_3}$ to model simple time-invariant information in the environment.

\subsubsection{Inverse Dynamics}
\label{sec:inverse_dynamics}
To enable disentanglement representation learning that corresponds to the control signals, we introduce the training objective of \textit{inverse dynamics}. 
This objective encourages the action-conditioned branch to learn a more deterministic state transition based on specific actions, while the action-free branch learns the remaining noncontrollable dynamics independent of the control signals.
Accordingly, we design an \textit{Inverse Cell} of a $2$-layer MLP to infer the actions that lead to certain transitions of the controllable states:
\begin{equation}
    \label{eq:inverse}
    \text{Inverse dynamics:} \quad \tilde{a}_{t-1}=\texttt{MLP}(s_{t-1}, s_t),
\end{equation}
where the inputs are the posterior representations in the action-conditioned branch.
By learning to regress the true behavior $a_{t-1}$, the Inverse Cell facilitates the action-conditioned branch to isolate the representation of the controllable dynamics.
We respectively use the prior state $\tilde{s}_t$ and the posterior state ${z}_t$ to generate the controllable visual component $\hat{o}_t^s \in \mathbb{R}^{3 \times H \times W}$ with mask $M^s_t \in \mathbb{R}^{1 \times H \times W}$ and the noncontrollable component $\hat{o}_t^z \in \mathbb{R}^{3 \times H \times W}$ with $M^z_t \in \mathbb{R}^{1 \times H \times W}$. 
By further integrating the static information extracted from the first $K$ frames, we have
\begin{equation}  
    \label{eq:combine_rl}
     \hat{o}_{t} = M_{t}^s \odot \hat{o}_t^s + M_{t}^z \odot \hat{o}_t^z +  (1-M_{t}^s - M_{t}^z) \odot \hat{o}^b,
\end{equation}
where $\hat{o}^b = \texttt{Dec}_{\varphi_3}(\texttt{Enc}_{\theta, \phi_3}(o_{1:K})))$.

For reward modeling, we have two options concerning the action-free branch. First, we may regard noncontrollable dynamics as irrelevant noises that do not contribute to the task and therefore do not involve $z_t$ in imagination.
In other words, the policy and predicted rewards would solely rely on controllable states, \textit{e.g.}, $p(r_t \mid s_t)$.  
Alternatively in other cases, we need to consider the influence of future noncontrollable states on the agent's decision-making process and incorporate the action-free components during behavior learning.
To achieve this, we train the reward predictor to model $p(r_t \mid s_t, z_t)$ in the form of MLPs.

For a training sequence of $(o_t, a_t, r_t)_{t=1}^T$ sampled from the replay buffer, the world model can be optimized using the following loss functions, where $\alpha$, $\beta_1$, and $\beta_2$ are hyper-parameters:
\begin{equation}
\label{eq:loss}
\begin{split}
\mathcal{L}_\text{base} = &\mathbb{E} \ \{
\sum_{t=1}^{T} \underbrace{-\ln p(o_{t} \mid h_{t}, s_{t}, h^\prime_t, z_{t})}_{\text {image log loss }}  +\underbrace{\alpha \ell_2(a_t, \tilde{a}_{t})}_{\text {action loss }}\\
&\underbrace{-\ln p(r_{t} \mid h_{t}, s_{t}, h^\prime_t, z_{t})}_{\text {reward log loss }} 
 \underbrace{- \ln p(\gamma_{t} \mid h_{t}, s_{t}, h^\prime_t, z_{t})}_{\text {discount log loss }}  \\ 
 &+\underbrace{\beta_1 \mathrm{KL}[q(s_{t} \mid h_{t}, o_{t}) \mid p(s_{t} \mid h_{t})]}_{\mathrm{KL} \text { divergence in the action-conditioned branch }} \\ &+\underbrace{\beta_2 \mathrm{KL}[q(z_{t} \mid h^\prime_{t}, o_{t}) \mid p(z_{t} \mid h^\prime_{t})]}_{\mathrm{KL} \text { divergence in the action-free branch }}\}.
\end{split}
\end{equation}

\vspace{-4pt}

\subsubsection{Training Collapse and Min-Max Variance Constraints}
\label{sec:variance}

\begin{figure}[t]
    \centering
    \includegraphics[width=0.9\linewidth]{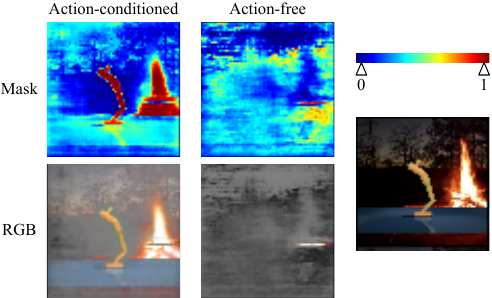}
    \caption{A showcase of training collapse that the action-conditioned branch in the original version of Iso-Dream dominates the learning process of both controllable and noncontrollable dynamics. The corresponding results of \model{} are shown in \fig{fig:dmc-visual}.}
    \label{fig:collapse}
    \vspace{-5pt}
\end{figure}

Despite modeling inverse dynamics, for the original Iso-Dream, we find that it is still challenging for the world model to isolate controllable and noncontrollable dynamics. 
We observed in the preliminary experiments that the disentanglement results were unstable over multiple runs of world model training, and that the action-conditioned branch occasionally learned mismatched representations of noncontrollable state transitions.
An example is shown in \fig{fig:collapse}.
It implies that most useful information may collapse into the action-conditioned branch while the action-free branch learns almost nothing, which we call ``\textit{training collapse}''.
This phenomenon arises due to the inherent limitations of the training objective in inverse dynamics, which may not always ensure the complete exclusion of action-independent state transitions, particularly when the action-conditioned network branch possesses a strong capacity for modeling dynamics.

To keep the state transition branches from training collapse, we propose the \textit{min-max variance constraints}, whose key idea is to (i) maximize the diversity of outcomes in the action-conditioned branch given distinct action inputs and (ii) minimize the diversity of outcomes in the action-free branch under similar conditions.
To this end, unlike in the original Iso-Dream, we make the action-free branch also aware of the action signal during the world model learning process. But for behavior learning and policy deployment, we simply set the input action to $0$-values.

There is an information-theoretic interpretation behind calculating variance. In order to investigate the connection between dynamics and action signals, the world model is enforced to identify the dynamics that provide information about our beliefs regarding the action signals. The expected information gain can be expressed as the conditional entropy of state and action:
\begin{equation}  
     I(s_{t}; a_{t-1}{\mid}s_{t-1}) = H(s_{t}{\mid}s_{t-1}) - H(s_{t}{\mid}s_{t-1}, a_{t-1}).
\end{equation}

As shown in \fig{fig:rl_model_and_policy} (right), for the action-conditioned branch, we maximize the mutual information between the state and action signal to focus on the state transition of a specific action.
Given a batch of hypothetical actions $\{a_{t-1}^i{\mid}i \in [1, n]\}$, for the same controllable state $s_{t-1}$, we have different state transitions based on these actions:
$\tilde{s}_t^i \sim p(\tilde{s}_t^i {\mid} s_{t-1}, a_{t-1}^i), \ i\in[1, n]$.
The empirical variance is used to approximate the information gain, and the objective can be written as
\begin{equation} 
\begin{split}
     L_{s} &= \max \sum\limits_t^{T} \mathrm{Var}(\tilde{s}_t^i) = \max \sum\limits_t^{T} \frac{1}{n-1} \sum\limits_i(\tilde{s}_t^i - \bar{s}_t)^2, \\
     \bar{s}_t &= \frac{1}{n} \sum\limits_i \tilde{s}_t^i, \quad  i\in[1, n].
\end{split}
\end{equation}
On the contrary, in the action-free branch, we minimize the variance of output states resulting from different actions, penalizing the diversity of state transitions:
\begin{equation} 
\begin{split}
     L_{z} &=  \min \sum\limits_t^{T} \mathrm{Var}(\tilde{z}_t^i) =  \min \sum\limits_t^{T} \frac{1}{n-1} \sum\limits_i(\tilde{z}_t^i - \bar{z}_t)^2, \\
     \bar{z}_t &= \frac{1}{n} \sum\limits_i \tilde{z}_t^i, \quad  i\in[1, n].
\end{split}
\end{equation}
The overall training objective of the world model is
\begin{equation} 
\begin{split}
\label{eq:all_loss}
     L_\text{all} = L_\text{base} + L_\text{var},
\end{split}
\end{equation}
where $L_\text{var} = \lambda_1 L_s + \lambda_2 L_z$. $\lambda_1$ and $\lambda_2$ are hyper-parameters.

For convenience, we only use two opposite actions $\{a_t, -a_t\}$ in the action-conditioned branch, and use the action set $\{a_t, 0, -a_t\}$ in the action-free branch to figure out $L_s$ and $L_z$. As for subsequent learning, we use $a_t$ and $0$ in the action-conditioned and action-free branches, respectively. 

\subsubsection{Sparse Dependency between Decoupled States}
\label{sec:spare-dependency}

\begin{figure}[t]
    \centering
    \includegraphics[width=0.85\linewidth]{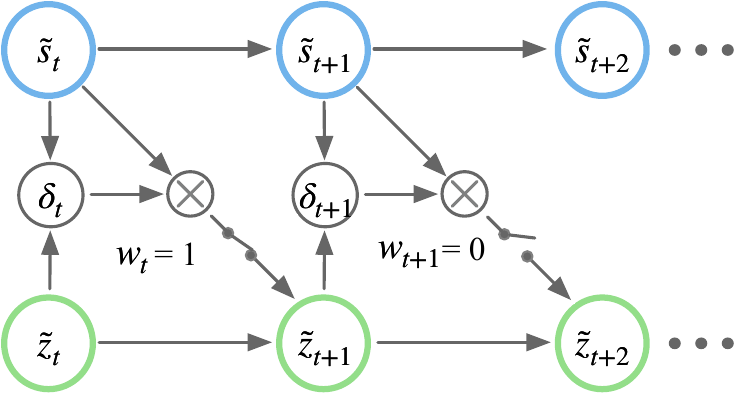}
    \caption{
    The dependency gate involves a binary gate that can either be open ($w_t=1$) or closed ($w_t=0$). When the gate is open, the transition of the next noncontrollable state $\tilde{z}_{t+1}$ takes into account the dependency between $\tilde{s}_t$ and $\tilde{z}_t$.}
    \label{fig:sparse_model}
    \vspace{-5pt}
\end{figure}

In certain situations, the controllable and noncontrollable dynamics are not entirely independent, as shown in \fig{fig:sparse_vis}. This is particularly true in autonomous driving, where the actions of the ego-agent can influence the behavior of other vehicles, causing them to steer or slow down. To accurately predict future noncontrollable states based on current controllable states, it is essential to account for these sparse dependencies.

To achieve effective modeling of sparse dependency, it is essential to identify the moment when controllable states exert a significant influence on noncontrollable states. 
In order to facilitate this, we present a compact module called the \textit{dependency gate}, which connects the previously isolated action-free and action-conditioned branches, as shown in \fig{fig:rl_model_and_policy} (left).
We unfold the detailed structure dependency gate in time (see \fig{fig:sparse_model}), where the controllable state $\tilde{s}_t$ and noncontrollable state $\tilde{z}_t$ are concatenated and passed through a fully connected layer represented by $f(\tilde{s}_t,\tilde{z}_t)$. A sigmoid function is then applied as an activation signal to control the gate, which is formulated as
\begin{equation}
\label{sparse_gate}
\delta_t(w_t=1{\mid}\tilde{s}_t, \tilde{z}_t)=\left\{
\begin{aligned}
1,& \quad \mathrm{sigmoid}(f(\tilde{s}_t, \tilde{z}_t)) \geq 0.5, \\
0,& \quad \mathrm{otherwise}.
\end{aligned}
\right.
\end{equation}
When the gate detects a dependency between controllable and noncontrollable states ($w_t = 1$), the subsequent noncontrollable state $\tilde{z}_{t+1}$ is determined by both $\tilde{s}_t$ and $\tilde{z}_t$ using the action-free transition, which is defined as follows: 
\begin{equation}
\label{eq:sparse}
\tilde{z}_{t+1} \sim p(\tilde{z}_{t+1} \mid \tilde{z}_t, w_t \odot \tilde{s}_t).
\end{equation}

\subsection{Behavior Learning in Isolated Imaginations}
\label{sec:behavior-learning}
\begin{figure}[t]
    \centering
    \includegraphics[width=0.9\linewidth]{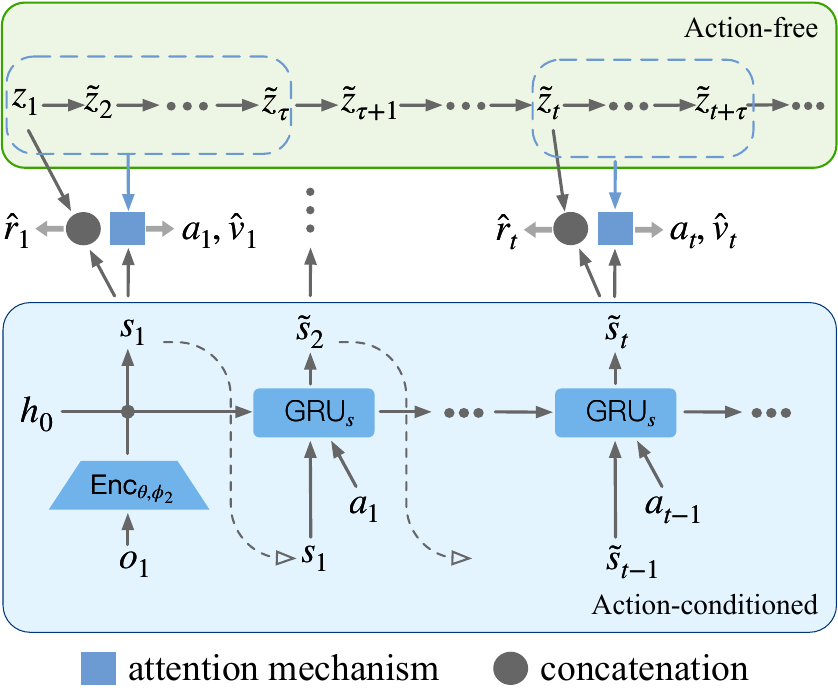}
    \caption{The agent learns \textit{future-dependent} policies in world model's imaginations through a \textit{future state attention} mechanism.
   }
    \label{fig:policy}
    \vspace{-5pt}
\end{figure}

Thanks to the decoupled world model, we can optimize agent behavior to adaptively consider the relationship between available actions and potential future states of the noncontrollable dynamics. 
A practical example is autonomous driving, where the movement of other vehicles can be naturally viewed as noncontrollable but predictable components. 
As shown in \fig{fig:policy}, we here propose an improved actor-critic learning algorithm that \textit{(i) allows the action-free branch to foresee the future ahead of the action-conditioned branch, and (ii) exploits the predicted future information of noncontrollable dynamics to make more forward-looking decisions}.

\begin{algorithm*}[t] 
  \caption{\model{} (Highlight: Our modifications to \textcolor{MyDarkBlue}{behavior learning} \& \textcolor{MyDarkGreen}{policy deployment} of the original Dreamer)}  
  \label{alg:algorithm}  
  \small
  \textbf{Hyper-parameters: }{$L$: Imagination horizon; $\tau$: Window size for future state attention}
  \begin{algorithmic}[1]
    \State Initialize the replay buffer $\mathcal{B}$ with random episodes. 
    \While{not converged}
    \For{update step $c = 1 \dots C$}
        \State \texttt{// Representation learning} 
        \State Draw data sequences $\left\{\left(o_{t}, a_{t}, r_{t}\right)\right\}_{t=1}^{T} \sim \mathcal{B}$. 
        \State Compute the controllable state $s_t \sim q(s_t {\mid} h_t, o_t, a_{t-1})$ and the noncontrollable state $z_t \sim q(z_t {\mid} h^{\prime}_t, o_t, a_{t-1})$. 
        \State Compute world model loss using Eq. \eqref{eq:all_loss} and update model parameters. 
        \State \texttt{// Behavior learning}  
        \For{time step $i = t \dots t+L$}
        \State \textcolor{MyDarkBlue}{Compute the next noncontrollable state $\tilde{z}_{i+1}$ using Eq. \eqref{eq:sparse}.}
        \State \textcolor{MyDarkBlue}{Roll-out the noncontrollable states $\{\tilde z_j\}_{j=i+2}^{i+\tau}$ from $\tilde{z}_{i+1}$ through the action-free branch alone.} 
        \State \textcolor{MyDarkBlue}{Compute latent state $e_i \sim \texttt{Attention}(\tilde s_i, \tilde z_{i: i+\tau})$ using Eq. \eqref{eq:attention}.}
        \State \textcolor{MyDarkBlue}{Imagine an action ${a}_i \sim \pi (a_i {\mid} e_i)$.} 
        \State \textcolor{MyDarkBlue}{Predict the next controllable state $\tilde s_{i+1} \sim p(\tilde s_{i}, {a}_{i})$ using the action-conditioned branch alone.}
        \EndFor
        \State Update the policy and value models in Eq. \eqref{eq:ac-model} using estimated rewards and values.
    \EndFor
    \State \texttt{// Environment interaction} 
    \State $o_{1} \leftarrow$ \texttt{env.reset}() 
    \For{time step $t = 1\dots T$}
    \State Calculate the posterior representation $s_{t} \sim q\left(s_{t} \mid h_t, o_{t}, a_{t-1}\right), z_{t} \sim q\left(z_{t} \mid h_t^\prime, o_{t}, a_{t-1}\right)$. 
    \State \textcolor{MyDarkGreen}{Compute the next noncontrollable state $\tilde{z}_{t+1} \sim p(\tilde{z}_{t+1}{\mid}z_{t}, w_t \odot s_{t})$ using Eq. \eqref{eq:sparse}.}
    \State \textcolor{MyDarkGreen}{Roll-out the noncontrollable states $\tilde{z}_{t+2:t+\tau}$ from $\tilde{z}_{t+1}$ through the action-free branch alone.}
    \State \textcolor{MyDarkGreen}{Generate $a_t \sim  \pi (a_t \mid s_t, z_t, \tilde{z}_{t+1:t+\tau})$ using future state attention in Eq. \eqref{eq:attention}.} 
    \State $r_t, o_{t+1} \leftarrow$ \texttt{env.step}($a_t$)
    \EndFor
    \State Add experience to the replay buffer $\mathcal{B} \leftarrow \mathcal{B} \cup\{\left(o_{t}, a_{t}, r_{t}\right)_{t=1}^{T}\}$.
    \EndWhile
  \end{algorithmic}  
\end{algorithm*}

Suppose we are making decisions at time step $t$ in the imagination period.
A straightforward solution from the original Dreamer method is to learn an action model and a value model based on the isolated controllable state $\tilde{s}_t \in \mathbb{R}^{1 \times d}$. 
With the aid of an attention mechanism, we can establish a connection between it and future noncontrollable states. It is important to note that we only employ sparse dependency in the initial imagination step to obtain $\tilde{z}_{t+1}$, as the subsequent controllable states are not yet available at time step $t$. Once we have predicted a sequence of future noncontrollable states $\tilde{z}_{t:t+\tau} \in \mathbb{R}^{\tau \times d}$, where $\tau$ is the sliding window length from the present time, we explicitly compute the relations between them using the following equation:
\begin{equation} 
\begin{split}
    \label{eq:attention}
     e_t = \mathrm{softmax}(\tilde s_t \ \tilde{z}^T_{t:t+\tau}) \ \tilde{z}_{t:t+\tau} + \tilde s_t.
\end{split}
\end{equation}
This equation allows us to dynamically adjust the horizon of future noncontrollable states using the attention mechanism.
In this way, $\tilde{s}_t$ evolves to a more ``\textit{visionary}'' representation $e_t \in \mathbb{R}^{1 \times d}$. We modify the action and value models in Dreamer \cite{hafner2019dream} as follows:
\begin{equation}
\label{eq:ac-model}
\begin{split}
    \text{Action model:}& \quad {a}_{t} \sim \pi(a_{t} \mid e_t ), \\
    \text{Value model:}& \quad v_{\xi}(e_{t}) \approx \mathbb{E}_{\pi\left(\cdot \mid e_{t}\right)} \sum_{k=t}^{t+L} \gamma^{k-t} r_{k},
\end{split}
\end{equation}
where $L$ is the imagination time horizon. 
As shown in \alg{alg:algorithm}, during imagination, we first use the action-free transition model to obtain sequences of noncontrollable states of length $L+\tau$, denoted by $\{\tilde{z}_i \}_{i=t}^{i+L+\tau}$.
At each time step in the imagination period, the agent draws an action ${a}_j$ from the visionary state $e_j$, which is derived from Eq. \eqref{eq:attention}. The action-conditioned branch uses the action ${a}_j$ in latent imagination and predicts the next controllable state $s_{j+1}$.
We follow DreamerV2 \citep{hafner2020mastering} to train our action model with the objective of maximizing the $\lambda$-return \citep{sutton2018reinforcement}, while our value model was trained to perform regression on the $\lambda$-return. For further information on the loss functions, please refer to Eq. (5-6) as detailed in the paper of DreamerV2 \citep{hafner2020mastering}.

\subsection{Policy Deployment by Rolling out Noncontrollable Dynamics}
\label{sec:Policy-Deployment}

During policy deployment, as shown in Lines 22-24 in Alg. \ref{alg:algorithm}, the action-free branch predicts the next-step noncontrollable states $\tilde{z}_{t+1}$ using Eq. \eqref{eq:sparse} and then consecutively rolls out the future noncontrollable states $\tilde{z}_{t+2:t+\tau}$ starting from $\tilde{z}_{t+1}$. 
Similar to Eq. \eqref{eq:attention} used in the process of behavior learning, the learned future state attention network is used to adaptively integrate $s_t$, $z_t$ and $\tilde{z}_{t+1:t+\tau}$. 
Based on the integrated feature $e_t$, the \model{} agent then draws $a_t$ from the action model to interact with the environment.
As discussed in \sect{sec:challenges}, if the noncontrollable dynamics are irrelevant to the control task, the policy at each time step $t$ is generated using only the state of controllable dynamics when interacting with the environment.

\section{Experiments}

\subsection{Experimental Setup}
\begin{figure}[t]
\vspace{-5pt}
    \centering
    \includegraphics[width=0.7\linewidth]{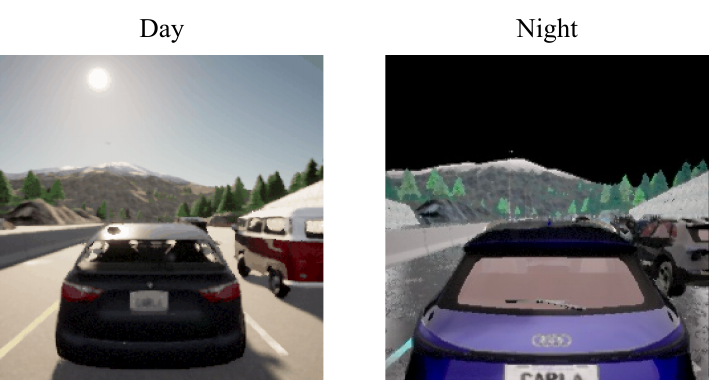}
    \vspace{-3pt}
    \caption{Examples of day and night modes in CARLA.}
    \label{fig:day_night_vis}
    \vspace{-12pt}
\end{figure}

\vspace{-5pt}
\myparagraph{Benchmarks.}
We evaluate \model{} on two RL environments:
\begin{itemize}[leftmargin=*]
\vspace{-3pt}
    \item
    \textbf{CARLA} \citep{DBLP:conf/corl/DosovitskiyRCLK17}: CARLA is a simulator with complex and realistic visual observations for autonomous driving research. We train our model to perform the task of first-person highway driving in ``Town04'', where the agent’s goal is to drive as far as possible in $1{,}000$ time steps without colliding with any of the $30$ other moving vehicles or barriers. In addition to our conference paper, we incorporate more diverse settings into our study, including both day and night modes as shown in \fig{fig:day_night_vis}.
    \item
    \textbf{DeepMind Control Suite} \citep{tassa2018deepmind}: 
    DMC contains a set of continuous control tasks and serves as a standard benchmark for vision-based RL. To evaluate the generalization of our method by disentangling different components under complex visual dynamics, we use two modified benchmarks \citep{hansen2021softda}, namely \texttt{video\_easy}, which contains $10$ simple videos, and \texttt{video\_hard}, which contains $100$ complex videos. 
\end{itemize}

\vspace{-5pt}
\myparagraph{Compared methods.}
We compare \model{} with the following visual RL approaches:
\begin{itemize}[leftmargin=*]
\vspace{-3pt}
    \item \textbf{DreamerV2 \citep{hafner2020mastering}}: A model-based RL method that learns directly from latent variables in world models. The latent representation allows agents to imagine thousands of trajectories in parallel.
    \item \textbf{DreamerV3 \citep{hafner2023mastering}}:
    A further improved version of Dreamer that learns to master diverse domains with fixed hyperparameters.
    \item \textbf{DreamerPro \citep{deng2022dreamerpro}}: A non-contrastive, reconstruction-free model-based RL method that combines Dreamer \citep{hafner2019dream} with prototypes to enhance robustness to distractions.
    \item \textbf{CURL \citep{srinivas2020curl}}: A model-free RL method that uses contrastive learning to extract high-level features from raw pixels, maximizing agreement between augmented data of the same observation.
    \item \textbf{SVEA \citep{hansen2021stabilizing}}: A framework for data augmentation in deep Q-learning algorithms that improves stability and generalization on off-policy RL.
    \item \textbf{SAC \citep{haarnoja2018soft}}: A model-free actor-critic method that optimizes a stochastic policy in an off-policy way.
    \item \textbf{DBC \citep{zhang2021learning}}: A method that learns a bisimulation metric representation without reconstruction loss. This representation is invariant to different task-irrelevant details in the observation. 
    \item \textbf{Denoised-MDP \citep{wang2022denoised}}: A framework that categorizes information out in the wild into four types based on controllability and relation with reward, and formulates useful information as that which is both controllable and reward-relevant.
\end{itemize}

\begin{figure*}[t] 
    \centering
	  \subfloat[]{
       \includegraphics[width=3.5in]{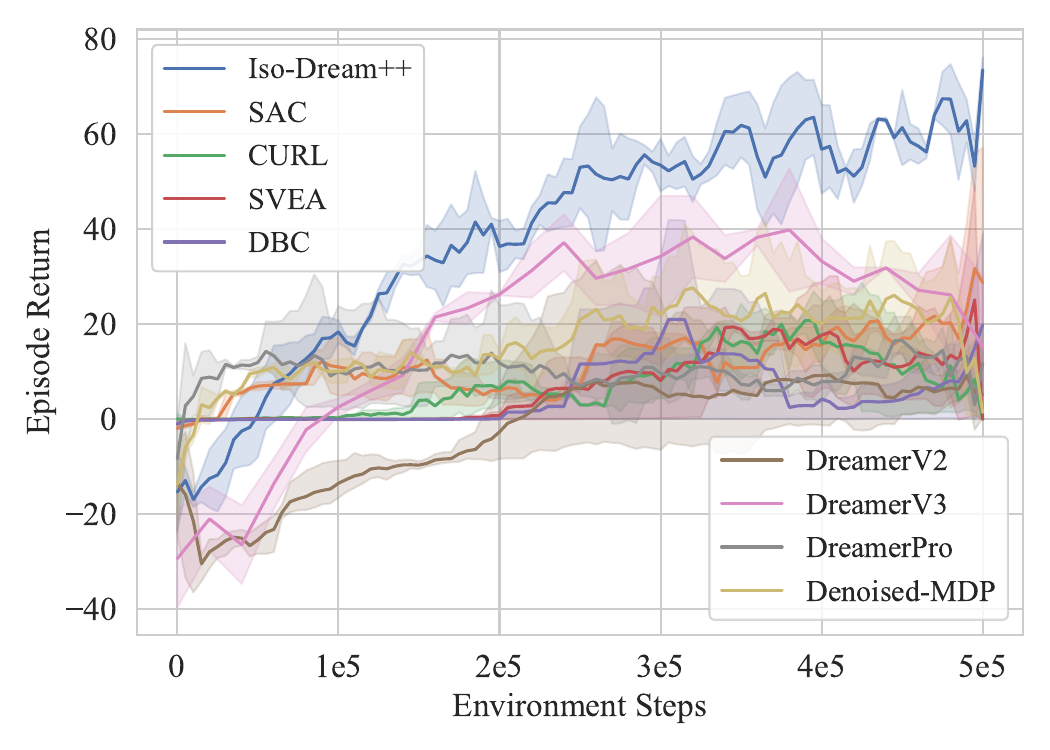}}
    \label{fig:compare_sota_carla}
    \hfil
	  \subfloat[]{
        \includegraphics[width=3.5in]{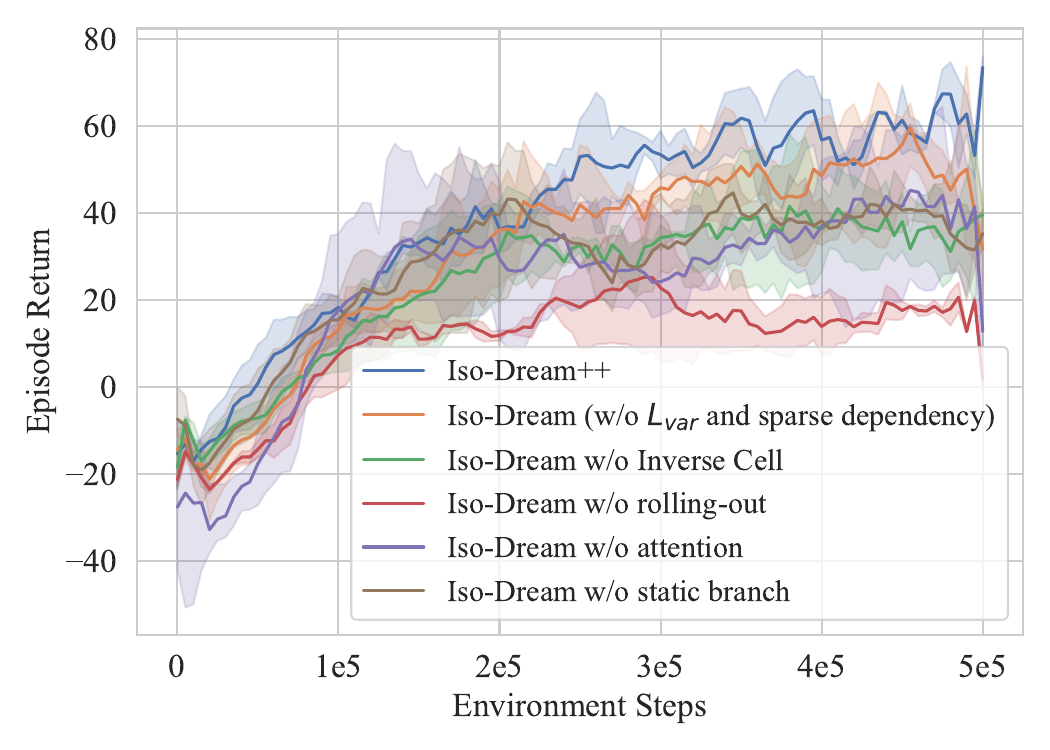}}
    \label{fig:ablation_carla_nips}
    \vspace{-5pt}
	  \caption{(a) Quantitative comparison with existing approaches for CARLA driving. (b) Ablation studies of individual effectiveness of inverse dynamics optimization (\textcolor{MyDarkGreen}{green}), noncontrollable rollouts (\textcolor{red}{red}), future-state attention (\textcolor{MyPurple}{purple}), and the separate branch for static information modeling (\textcolor{brown}{brown}). We also compare \model{} with its predecessor in our conference paper (\textcolor{orange}{orange}).}
	  \label{fig:compare_carla}
   \vspace{-5pt}
\end{figure*}

\begin{figure*}[!t]
    \centering
    \includegraphics[width=\linewidth]{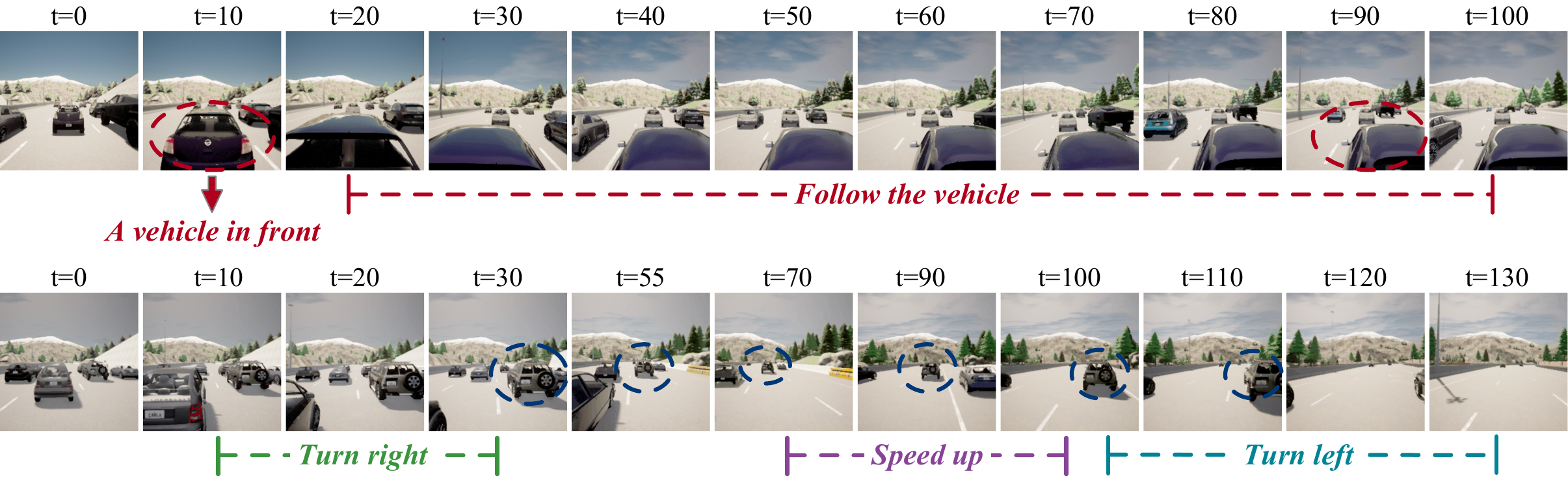}
    \vspace{-18pt}
    \caption{Examples of \model{} without (\textbf{top}) and with (\textbf{bottom}) sparse dependency. The agent without sparse dependency tends to follow the vehicle in front of it when there are many vehicles in the way. In the bottom row, the agent overtakes and accelerates flexibly.}
    \label{fig:sparse_ablation_vis}
    \vspace{-5pt}
\end{figure*}

\subsection{CARLA Autonomous Driving Environment}

\vspace{-5pt}
\myparagraph{Implementation.}
In the autonomous driving task, We use a camera with a $60$ degree view on the roof of the ego-vehicle, which obtains images of $64 \times 64$ pixels. Following the setting in the DBC \citep{zhang2021learning}, in order to encourage highway progression and penalize collisions, the reward is formulated as $r_t=v^T_{ego}\hat{u}_h \cdot \Delta t -\xi_1 \cdot \mathbbm{1} - \xi_2 \cdot |steer| $, where $v_{ego}$ is the velocity vector of the ego-vehicle, projected onto the highway’s unit vector $\hat{u}_h$, and multiplied by time discretization $\Delta t = 0.05$ to measure highway progression in meters. We use $\mathbbm{1} \in \mathbb{R^+}$ for collisions and a steering penalty $steer \in [-1,1]$ to facilitate lane-keeping. The hyper-parameters are set to $\xi_1=10^{-4}$ and $\xi_2=1$, respectively. We use $\beta_1 = \beta_2 = 1$ and $\alpha = 1$ in Eq. \eqref{eq:loss}, $\lambda_1 = \lambda_2 = 1$ in \eqn{eq:all_loss}, and $\tau=5$ in Eq. \eqref{eq:attention}.

\myparagraph{Quantitative comparisons.}
We present the quantitative results in CARLA in \fig{fig:compare_carla}(a). 
\model{} outperforms the compared models, including DreamerV2, DreamerV3, DreamerPro, and Denoised-MDP, significantly. 
After $500k$ environment steps, \model{} achieves an average return of around $60$, while DreamerV2 and Denoised-MDP achieve $10$ and $25$ respectively. 
In DreamerV2, the latent representations contain both controllable and noncontrollable dynamics, which increases the complexity of modeling the state transitions in imagination. 
Compared with Denoised-MDP, which also decouples information according to controllability, \model{} has the advantage of making forward-looking decisions by rolling out future noncontrollable states.

\myparagraph{Ablation studies.}
\fig{fig:compare_carla}(b) provides the ablation study results that validate the effectiveness of inverse dynamics, the rolling-out strategy of noncontrollable states, the attention mechanism, and the modeling of static information.
As shown by the green curve, removing the Inverse Cell reduces the performance of \model{}, which indicates the importance of isolating controllable and noncontrollable components.
For the model indicated by the red curve, we do not use the rollouts of future noncontrollable states as the inputs of the action model. The results demonstrate that rolling out noncontrollable states in the action-free branch significantly improves the agent's decision-making results by perceiving potential risks in advance.
Moreover, we evaluate \model{} without attention mechanism where the action model directly concatenates the current controllable state with a sequence of future noncontrollable states and takes them as inputs. As shown by the purple curve, the attention mechanism extracts valuable information from future noncontrollable dynamics better than concatenation.
Furthermore, as shown by the brown curve, our approach’s performance decreases by about $15\%$ without a separate network branch for capturing the static information.
%
Moreover, a comparison between the blue curve and the orange curve reveals a decline in our model's performance when we remove the min-max variance constraints and sparse dependency modeling. 
Unlike the DMC suite, where the original Iso-Dream is more vulnerable to training collapse, in the CARLA environment, the sparse dependency modeling method plays a crucial role in the improved performance of \model{}.
In \fig{fig:sparse_ablation_vis}, we present visual examples produced by our models with and without the sparse dependency.
Without sparse dependency (top row), the agent fails to predict that other vehicles will slow down or brake when changing lanes, making it safer to follow the vehicle ahead rather than overtake it during traffic congestion.
However, as shown in the bottom row of \fig{fig:sparse_ablation_vis}, the agent can decide whether to overtake or not based on its surroundings. 
These results indicate that sparse dependency greatly models the situation that the noncontrollable dynamics are affected by the controllable dynamics, which is conducive to the downstream decision-making task by accurately predicting the noncontrollable dynamics at future moments.

\begin{figure*}[t]
\begin{center}
\vspace{-5pt}
\centerline{\includegraphics[width=0.99\linewidth]{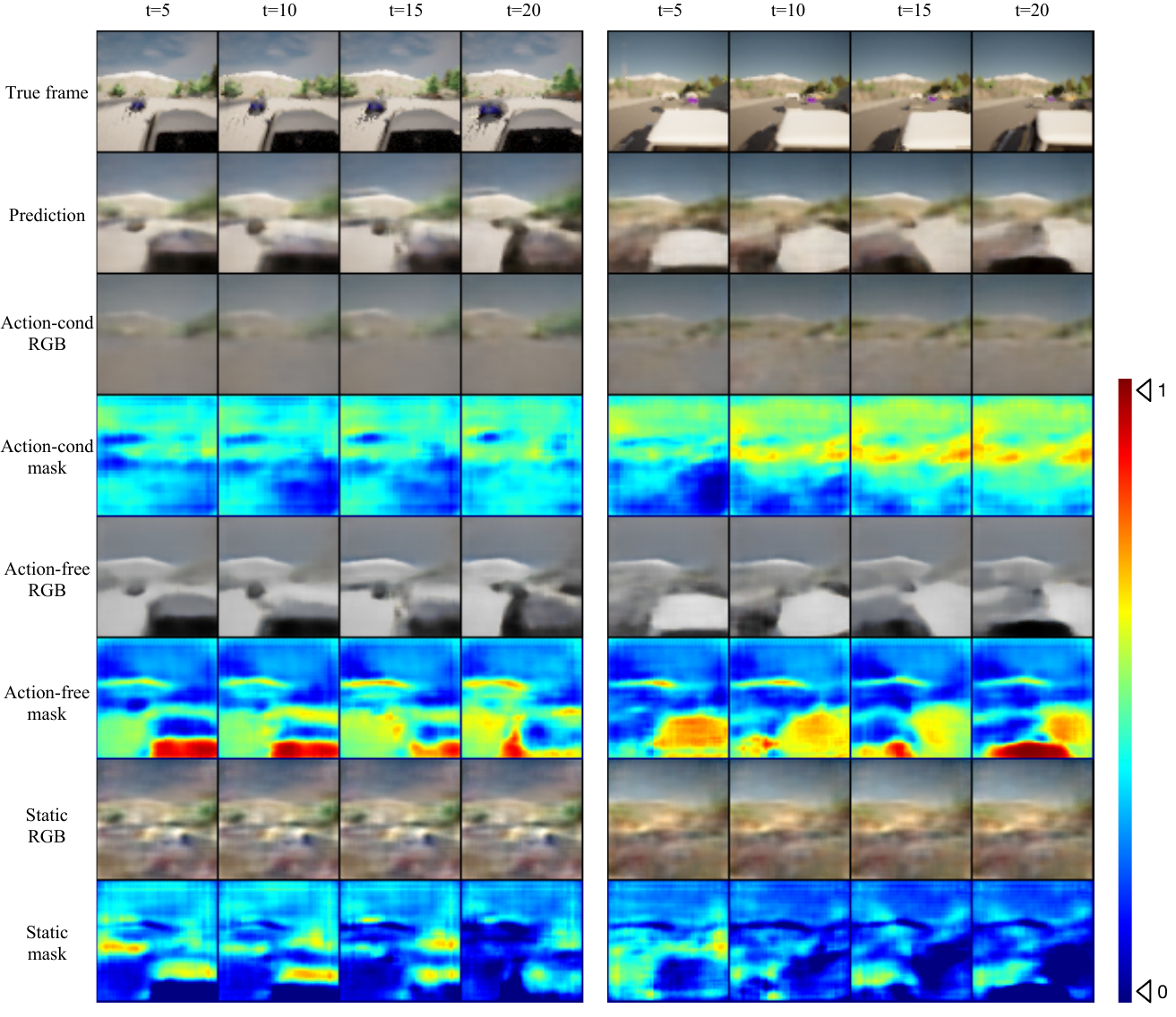}}
\vspace{-10pt}
\caption{Video prediction results in CARLA. For each sequence, we use the first $5$ images as context frames. The visual decoupled components (Rows 3, 5, 7) and masks (Rows 4, 6, 8) of each branch are presented. \model{} successfully isolates noncontrollable dynamics from the complicated environment, \textit{i.e.}, other driving vehicles.}
\label{fig:carla-visual}
\end{center}
\vspace{-25pt}
\end{figure*}

\begin{figure}[t]
\vspace{-5pt}
    \centering
    \includegraphics[width=\linewidth]{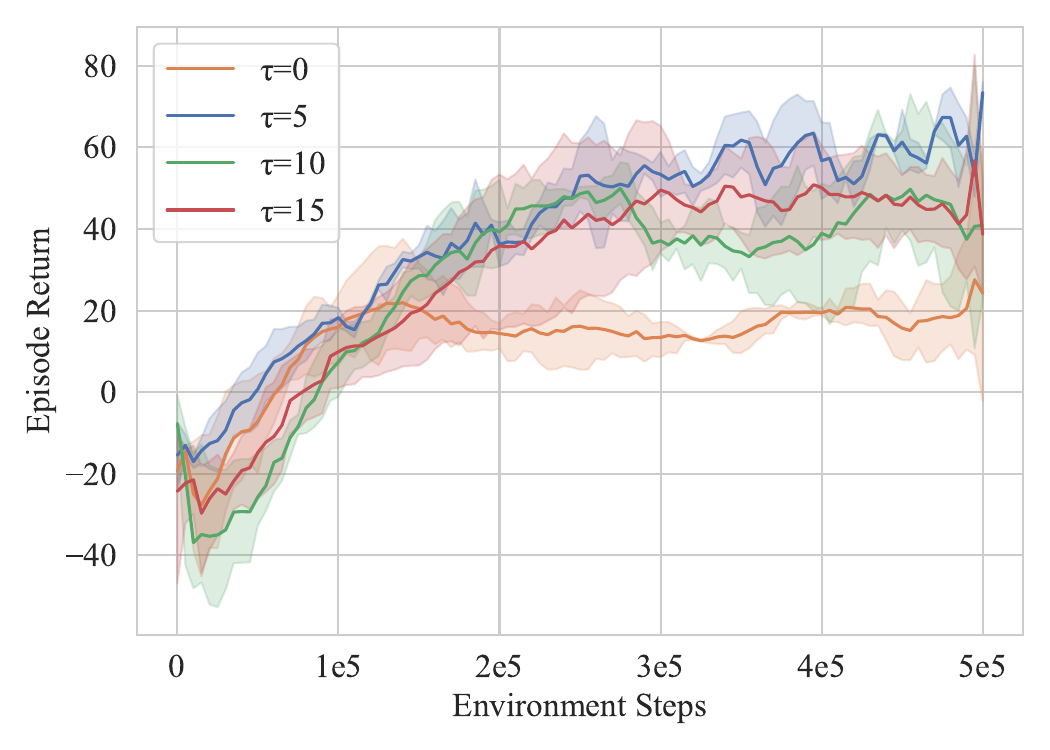}
    \vspace{-22pt}
    \caption{The hyperparameter analysis of $\tau$ in CARLA.}
    \label{fig:different_tau}
    \vspace{-18pt}
\end{figure}

\vspace{-2pt}
\myparagraph{Qualitative results.}
We show the video prediction results of \model{} in the CARLA environment in \fig{fig:carla-visual}. 
Because of the first-person view in this environment, the agent actions potentially affect all pixel values in the observation, as the camera on the main car (\textit{i.e.}, the agent) moves. Therefore, we can view the dynamics of other vehicles as a combination of controllable and non-controllable states. 
Accordingly, our model determines which component is dominant by learning attention mask values between $0$ and $1$ across the action-conditioned and action-free branches. The ``action-free masks'' present hot spots around other vehicles, while the attention values in corresponding areas on the ``action-cond masks'' are still greater than zero.
As shown in the third and fifth lines, \model{} mainly learns the dynamics of mountains and trees in the action-conditioned branch and the dynamics of other driving vehicles in the action-free branch, respectively, which helps the agent avoid collisions by rolling out noncontrollable components to preview possible future states of other vehicles.

\myparagraph{Hyperparameter analyses of $\tau$.}
We evaluate the effect of using different numbers of rollout steps of the noncontrollable states as the inputs of the action model. 
From the results in \fig{fig:different_tau}, we observe that our model achieves the best performance at $\tau=5$.
However, there are no remarkable differences among $\tau \in [5,10,15]$, as long-term predictions for noncontrollable states may increase model errors. 
Besides, we implement a model without rolling out noncontrollable states into the future, \textit{i.e.}, $\tau=0$. It performs significantly worse than other baselines with $\tau \in [5,10,15]$, which demonstrates the benefit of rolling out the disentangled action-free branch in policy optimization.

\begin{figure*}[!t]
    \centering
    \includegraphics[width=\linewidth]{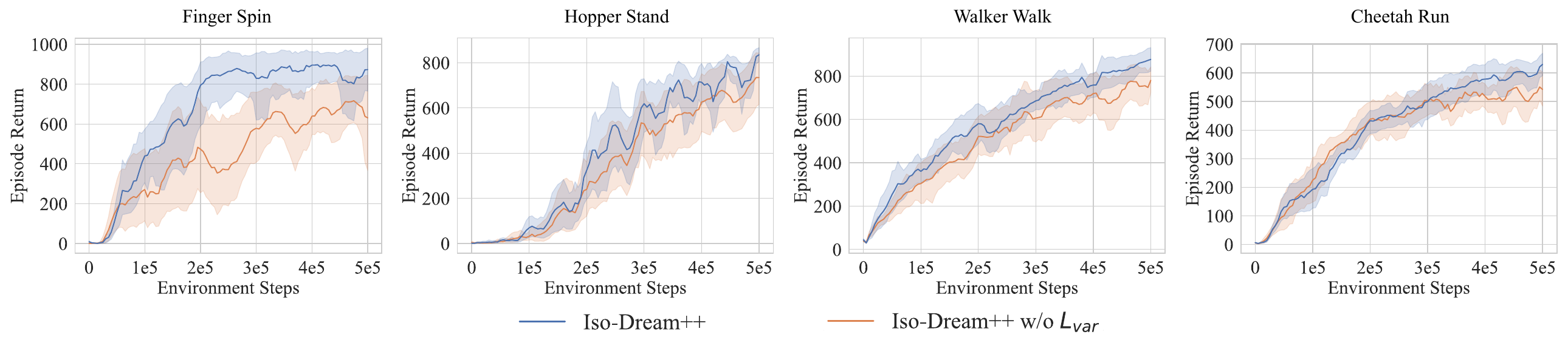}
    \vspace{-18pt}
    \caption{The ablation study of the proposed variance constraints in DMC. We report the results averaged over $10$ seeds.}
    \vspace{-10pt}
    \label{fig:ablation_variance_dmc}
\end{figure*}

\begin{table}[t]
  \centering
  \vspace{3pt}
  \caption{Qualitative results in DMC. The agents are trained and evaluated in environments with \texttt{video\_easy} background. * indicates a different setup from that of DBC. Iso-Dream (\textit{conf.}) is the model from our conference paper, which only uses the reconstruction loss (w/o KL divergence) in the action-free branch.}
  \label{tab:dmc_result}
  \vspace{-10pt}
    \setlength\tabcolsep{4.5pt}
    \begin{center}
    \begin{tabular}{l|ccccccc}
    \toprule    
    \multirow{2}{*}{Method}  & Finger & Hopper &  Walker  & Cheetah   \\
    & Spin & Stand & Walk & Run \\
    \midrule
    SVEA  & 562 $\pm$ 22 & 6 $\pm$ 8 & 826 $\pm$ 65 & 178 $\pm$ 64 \\
    CURL  & 280 $\pm$ 50 & 451 $\pm$ 250 & 443 $\pm$ 206 & 269 $\pm$ 24\\
    DBC*  & 1 $\pm$ 2 & 5 $\pm$ 9 & 32 $\pm$ 7 & 15 $\pm$ 5 \\
    DreamerV2   & 755 $\pm$ 92 & 260 $\pm$ 366 & 655 $\pm$ 47 & 475 $\pm$ 159 \\
    DreamerV3 & 124 $\pm$ 52 & 472 $\pm$ 328 & 701 $\pm$ 114 & 546 $\pm$ 117 \\
    DreamerPro   & 721 $\pm$ 147 & 295 $\pm$ 129 & 813 $\pm$ 88 & 297 $\pm$ 63 \\
    Denoised-MDP  & 635 $\pm$ 284 & 104 $\pm$ 117 & 214 $\pm$ 56 & 233 $\pm$ 119 \\
    \midrule
    Iso-Dream (\textit{conf.})  & 800 $\pm$ 59 & 746 $\pm$ 312 & 911 $\pm$ 50 & \textbf{659 $\pm$ 62} \\
    Iso-Dream & 816 $\pm$ 16 & 769 $\pm$ 173 & 852 $\pm$ 97 & 597 $\pm$ 156  \\
    Iso-Dream++   &\textbf{ 938 $\pm$ 51} & \textbf{877 $\pm$ 34} & \textbf{932 $\pm$ 37} & \textbf{639 $\pm$ 19} \\
    \bottomrule
    \end{tabular}
\end{center}
\vspace{-12pt}
\end{table}

\subsection{DeepMind Control Suite}
\label{expri:DMC}

\vspace{-3pt}
\myparagraph{Implementation.}
We evaluate our model on \texttt{video\_easy} and \texttt{video\_hard} benchmarks from the DMC Generalization Benchmark~\citep{hansen2021softda}, where the background is continually changing throughout an episode. 
All experiments use visual observations only, of shape $64 \times 64 \times 3$. The episodes last for $1{,}000$ steps and have randomized initial states. We apply a fixed action repeat of $R = 2$ across tasks. 
%
In this environment, since the background is randomly replaced by a real-world video, the noncontrollable motion of the background will affect the procedure of dynamics learning and behavior learning.
Therefore, to obtain a better decision policy and avoid the disruption from noisy backgrounds, the agent may decouple noncontrollable representation (\textit{i.e.}, dynamic background) and controllable representation in spacetime, and only use controllable representation for control, thereby removing the modeling sparse dependency.
Instead of training the action-free branch with only reconstruction loss in our preliminary work \citep{paniso}, we follow the structure described in \sect{sec:wm} since the noncontrollable dynamics in some video backgrounds are complicated for learning, particularly \texttt{video\_hard} benchmark.
We evaluate the models using $4$ tasks, \textit{i.e.}, Finger Spin, Cheetah Run, Walker Walk, and Hopper Stand. The maximum number of environmental steps is $500k$. We use $\beta_1 = \beta_2 = 1$ and $\alpha = 1$ in Eq. \eqref{eq:loss} and $\lambda_1 = \lambda_2 = 1$ in Eq. \eqref{eq:all_loss}.

\begin{figure}[t]
    \centering
    \includegraphics[width=\linewidth]{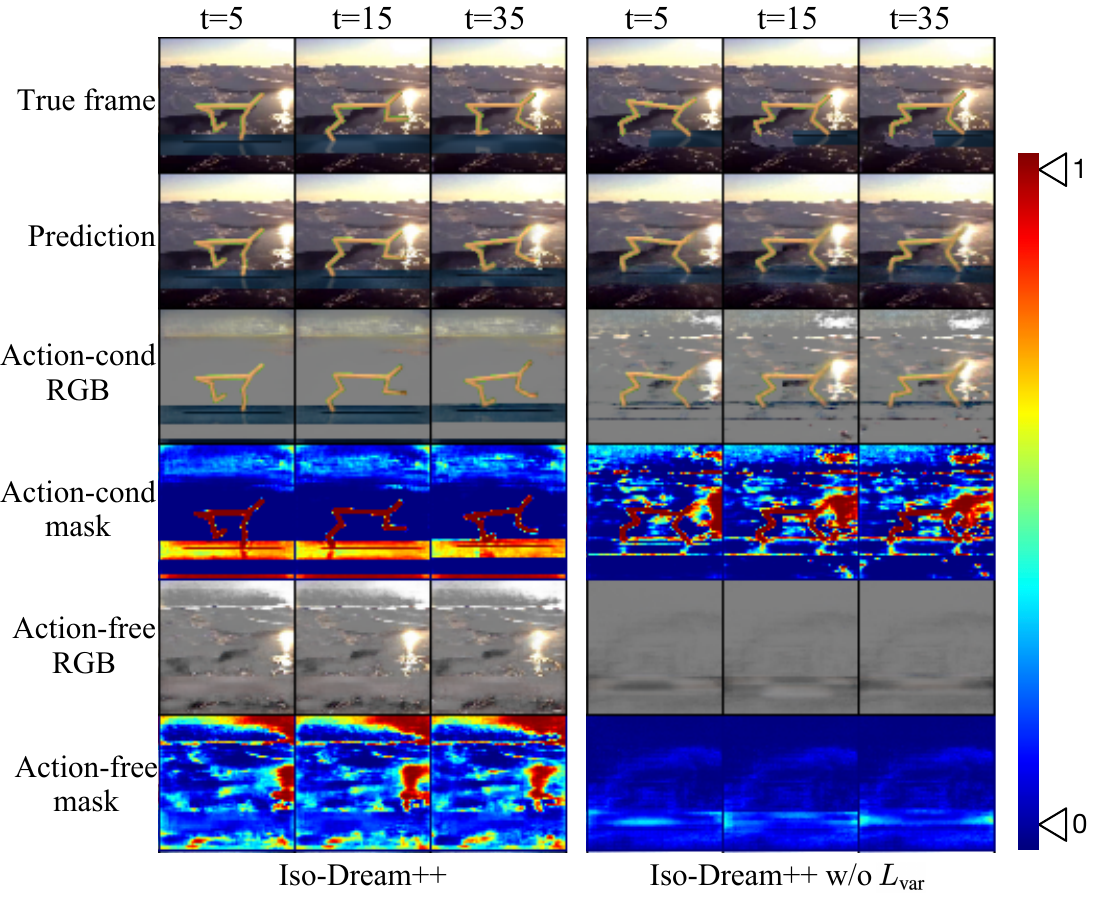}
    \vspace{-20pt}
    \caption{Video prediction results from our approaches w/ and w/o the proposed variance constraints.
   }
    \label{fig:ablation_variance_dmc_vis}
    \vspace{-12pt}
\end{figure}

\begin{figure*}[!t]
\begin{center}
\centerline{\includegraphics[width=0.99\linewidth]{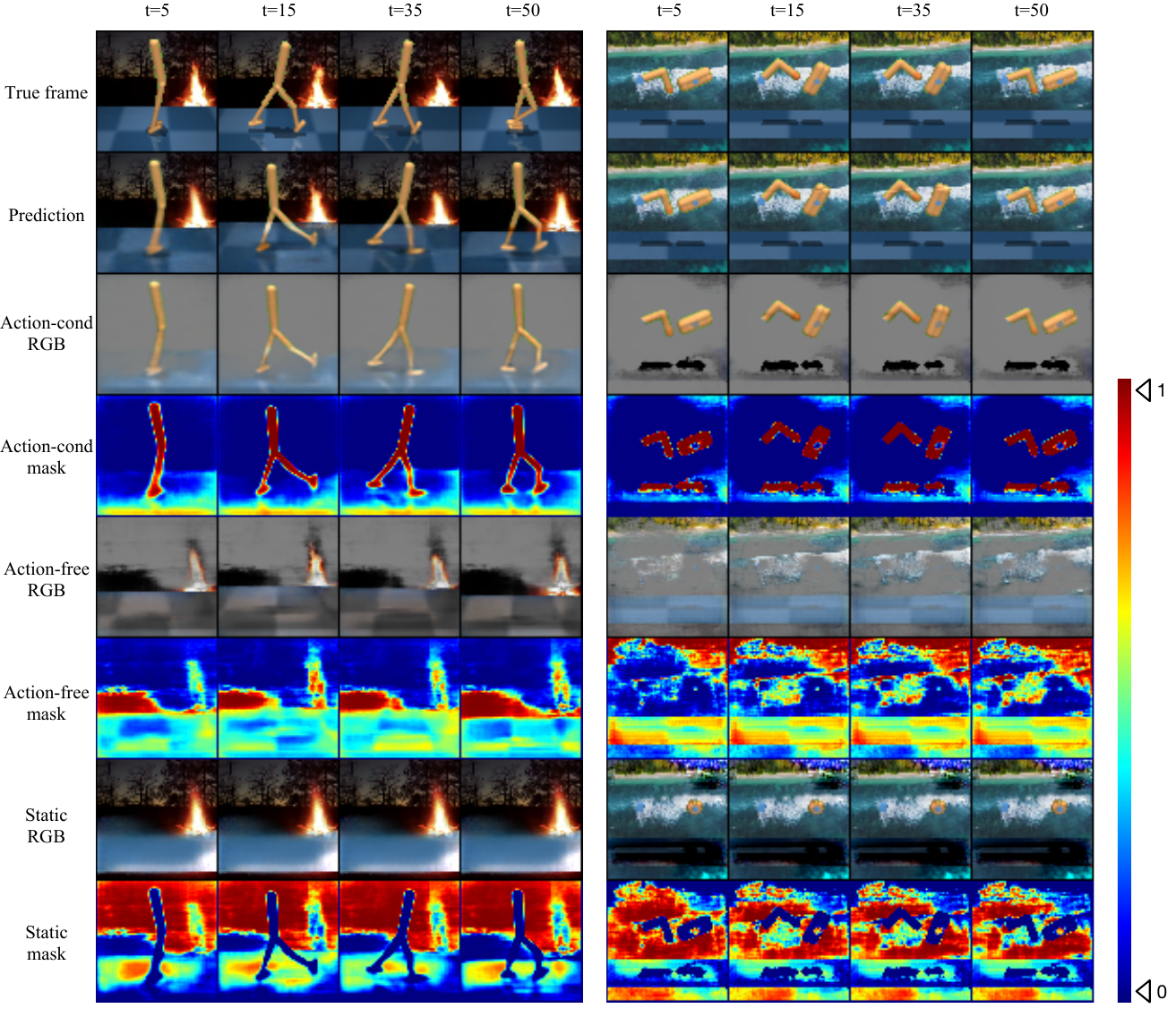}}
\vspace{-10pt}
\caption{Video prediction results with noisy backgrounds in DMC. For each sequence, we use the first $5$ images as context frames. \model{} successfully disentangles controllable and noncontrollable components.}
\label{fig:dmc-visual}
\end{center}
\vspace{-20pt}
\end{figure*}

\myparagraph{Quantitative comparisons.} 
We present the quantitative results of \model{} for \texttt{video\_easy} benchmark in \tab{tab:dmc_result}. 
Our final model outperforms DreamerV2 and other baselines significantly in all tasks. 
Compared with DBC and Denoised-MDP, which both aim to extract task-relevant representation from complex visual distractions, our method is more powerful with large performance gains on all four tasks, indicating that disentangling different dynamics by modular structure and variance constraints provides more cleaner and useful information for the downstream task.
Moreover, we have a better performance than DreamerPro, which is also based on Dreamer but learns the world model without reconstructing the observations. 
This demonstrates that our model effectively helps the agent to learn controllable visual representations and alleviate complex background interference.

\myparagraph{Analyses of the min-max variance constraints.}
We investigate the effectiveness of the proposed variance constraints described in \sect{sec:variance} by removing it from the training process of \model{}.  
As shown in \fig{fig:ablation_variance_dmc}, the compared models are trained for $10$ seeds, and the proposed method improves the performance of our model in most tasks, especially in \textit{finger spin}, where we have witnessed significant training collapse (see \fig{fig:collapse}).
In \fig{fig:ablation_variance_dmc_vis}, we provide a qualitative comparison of the disentanglement results between models trained with and without variance constraints.
Comparing the fifth and sixth row of action-free branch outputs, we observe that the action-free dynamics (such as the light over the lake) are correctly assigned to the action-free branch by variance constraints, preventing the action-conditioned branch from capturing all dynamic information, \textit{i.e.}, training collapse. Because of the pure dynamics captured in the action-conditioned branch, our model with variance constraints gains definite improvements.

\begin{table}[t]
  \centering
  \vspace{3pt}
  \caption{The study of the generalization ability and robustness of the compared models to immediate visual distractions in DMC.}
  \label{tab:resistance}
    \vspace{-10pt}
    \begin{center}
    \begin{tabular}{l|cccc}
    \toprule    
    \multirow{2}{*}{Method}  & Finger & Hopper &  Walker  & Cheetah   \\
    & Spin & Stand & Walk & Run \\
    \midrule
    \multicolumn{5}{l}{Train: \textit{video\_easy}; Test: \textit{video\_hard}} \\
    \midrule
    DreamerPro   & 628 $\pm$  151 & 180 $\pm$ 96 & 533 $\pm$ 212 &   244 $\pm$  27 \\
    Denoised-MDP   & 27   $\pm$  21 & 44  $\pm$   25 & 169  $\pm$   61 & 103 $\pm$ 46 \\
    Iso-Dream++  & \textbf{692 $\pm$ 185}  & \textbf{643 $\pm$  155} & \textbf{642 $\pm$ 129} & \textbf{441 $\pm$ 183} \\
    \midrule
    \multicolumn{5}{l}{Train: \textit{video\_easy}; Test: \textit{video\_easy with Gaussian noises}} \\
    \midrule
    DreamerPro   & 663 $\pm$  129 & 223   $\pm$  76 & 824 $\pm$  72 &  263 $\pm$ 55 \\
    Denoised-MDP   & 652  $\pm$ 306 & 103  $\pm$  110 & 180  $\pm$ 79 & 195 $\pm$  131  \\
    Iso-Dream++   & \textbf{851 $\pm$ 109} & \textbf{806 $\pm$  74} &  \textbf{906 $\pm$ 31} & \textbf{582 $\pm$ 69} \\
    \bottomrule
    \end{tabular}
\end{center}
\vspace{-18pt}
\end{table}

\myparagraph{Qualitative results.}
We use \model{} to perform video prediction in the DMC environment with \texttt{video\_easy} backgrounds. The frame sequence and actions are randomly collected from test episodes. The first $5$ frames are given to the model and the next $45$ frames are predicted only based on action inputs. 
In this environment, the video background can be viewed as a combination of noncontrollable dynamics and static representations.
\fig{fig:dmc-visual} visualizes the entire generated RGB images, the decoupled RGB components, and the corresponding masks of the three network branches. 
From these results, we observe that our approach has the ability to predict long-term sequence and disentangle controllable (agent) and noncontrollable dynamics (background motion) from complex visual images. As shown in the third and fourth rows in \fig{fig:dmc-visual}, the controllable representation has been successfully isolated and matches its mask. 
As shown in the fifth and sixth rows, the motion of fires and sea waves are captured as noncontrollable dynamics by the action-free branch.

\myparagraph{Robustness to immediate distractions.}
To assess the ability of \model{} to resist immediate visual distractions, we train the RL models on the \textit{video\_easy} benchmark and evaluate them on (i) \textit{video\_hard}; (ii) \textit{video\_easy with Gaussian noises}.
In Table \ref{tab:resistance}, we compare the results from \model{} with those from DreamerPro and Denoised-MDP, which both focus on learning robust representations against visual noises.
We observe a remarkable advantage of \model{} against unexpected distractions, which consistently outperforms the DreamerPro and Denoised-MDP across all tasks.

\begin{figure}[t]
    \centering
    \includegraphics[width=\linewidth]{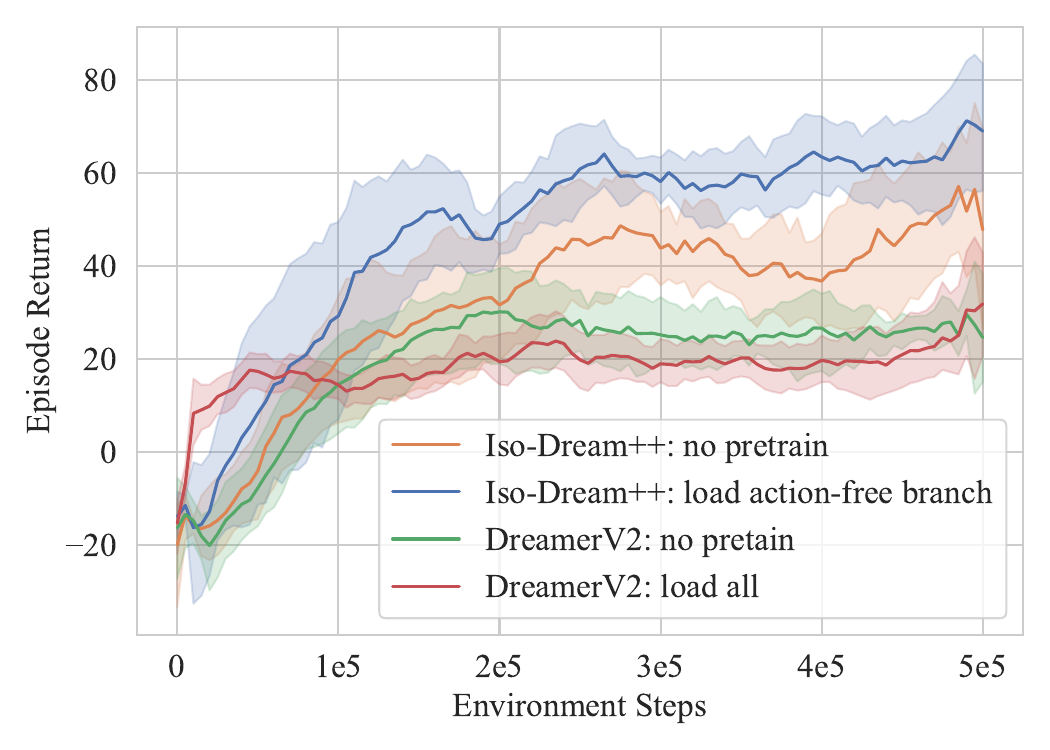}
    \vspace{-22pt}
    \caption{Transfer learning results across \texttt{Day} and \texttt{Night} modes in DMC. Leveraging a pretrained \textit{Day-mode} action-free branch can greatly benefit the finetuning results of \model{} in the \textit{Night mode}. Results are averaged over $10$ seeds.}
    \label{fig:day_to_night}
    \vspace{-8pt}
\end{figure}

\begin{figure*}[!b]
    \centering
    \vspace{-10pt}
    \includegraphics[width=\linewidth]{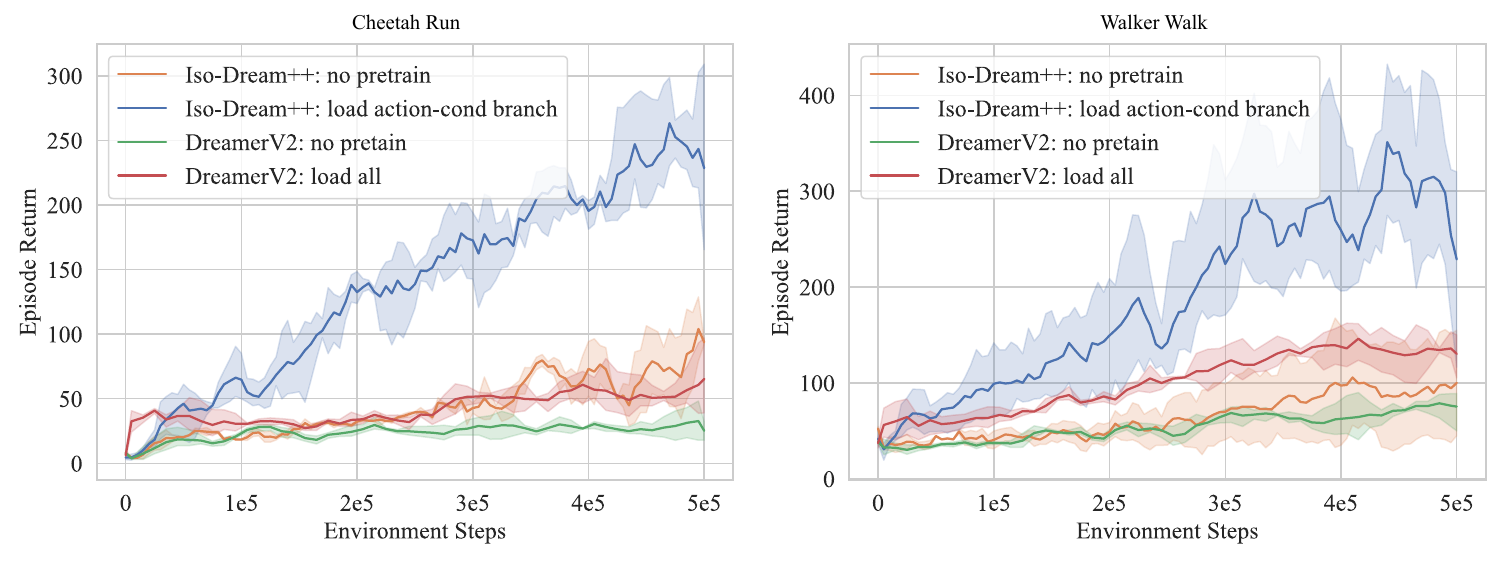}
    \vspace{-22pt}
    \caption{Transfer learning results in DMC across environments with \texttt{video\_easy} and \texttt{video\_hard} backgrounds. 
    Unlike DreamerV2, which can only transfer the entire world model (\textcolor{red}{red curve}) to the target domain, \model{} enables us to separately transfer the pretrained action-conditioned branch and obtain significantly better finetuning results (\textcolor{blue}{blue curve}). }
    \label{fig:easy_to_hard_two_tasks}
\end{figure*}

\subsection{Transfer Learning Analyses}
\vspace{-3pt}
\myparagraph{Transfer of noncontrollable dynamics in CARLA.}
Our model learns different dynamics in different branches, which makes it naturally suitable for transfer learning. Unlike common methods that transfer all knowledge from a pretrained source task, we can selectively transfer specific knowledge for a target task. 
Specifically, we only transfer relevant knowledge, such as shared dynamics between source and target tasks, to achieve precise disentanglement and robust decision-making on the target task.
In \fig{fig:day_night_vis}, We can see that the noncontrollable dynamics are similar between day and night modes, \textit{i.e.}, the movement of other driving vehicles.
We keep the action-free branch pretrained on the day mode of the CARLA environment and then train it on the night mode. The results are shown in \fig{fig:day_to_night}.
Comparing the orange curve and the blue curve, our model that transfers noncontrollable dynamics in the action-free branch has a significant improvement. However, the performance gain of DreamerV2 is small. 
Therefore, due to the modular structure in our \model{}, when there are two environments with similar dynamics, we can train on the easy environment first, and then load the specific pretrained branch to help the model learn on difficult tasks.
Therefore, the modular structure of our \model{} allows us to selectively transfer controllable or noncontrollable parts to novel domains based on our prior knowledge of the domain gap.

\myparagraph{Transfer of controllable dynamics in DMC.}
For DMC, we use the \texttt{video\_easy} (source) and \texttt{video\_hard} (target) benchmarks to evaluate the transfer ability of our model.
We transfer the controllable information in the action-conditioned branch because the controllable dynamics are the same in both environments, \textit{i.e.}, the motion of the agent.
From
\fig{fig:easy_to_hard_two_tasks}, we have two key observations. 
First, upon loading the pretrained action-conditioned branch, \model{} exhibits a significant advantage over the non-pretrained counterpart.
Second, it is noteworthy that the performance improvement achieved by our model through pretraining surpasses that of DreamerV2 by a considerable margin.

\section{Related Work}

\vspace{-3pt}
\myparagraph{Visual RL.}
In visual control tasks, RL agents learn the action policy directly from high-dimensional observations.
They can be roughly grouped into two categories, that is, model-free methods \citep{haarnoja2018soft,srinivas2020curl,yarats2019improving,kostrikov2020image,laskin2020reinforcement,hansen2021stabilizing} and model-based methods \citep{finn2017deep,oh2017value,ha2018world,hafner2019learning,hafner2019dream,kaiser2020model,sekar2020planning,zhang2021learning,bharadhwaj2022information,wang2022denoised,deng2022dreamerpro,xu2022RewardFree,ji2022update}.
%
%
Among them, the MBRL approaches explicitly model the state transitions and generally yield higher sample efficiency than the model-free methods.
A notable branch of work is the MuZero models, such as Stochastic MuZero \citep{antonoglou2022planning}. These models simulate and explore possible future action trajectories through Monte Carlo tree search (MCTS), which can effectively improve long-term decision-making but introduce a vast computational cost.
Notably, our model is different from Stochastic MuZero in two ways. First, we improve dynamics learning by encouraging representation decoupling, which we assume can enable the model to better understand the controllable and noncontrollable parts of the environment and greatly benefit the learned policy.
Second, unlike in Stochastic MuZero, our model performs an actor-critic algorithm without MCTS, which is practical in short-term control tasks such as autonomous driving and ensures higher efficiency for both policy optimization and deployment.
Another line of work is the so-called \textit{World Models}. 
Ha and Schmidhuber \cite{ha2018world} proposed to learn compressed latent states of the environment in a self-supervised manner and optimize potential behaviors based on the latent states generated by the world model.
Similarly, PlaNet \citep{hafner2019learning} introduces the \textit{recurrent state-space model} (RSSM) as the world model and performs the cross-entropy method over the imagined recurrent states. 
DreamerV1-V3 \citep{hafner2019dream,hafner2020mastering,hafner2023mastering} employ actor-critic methods to optimize the expected values and agent's behaviors over the predicted latent states in RSSM.
Specifically, our model based on DreamerV2 outperforms DreamerV2-V3 remarkably in CARLA and DMC. We also note that the state decoupling and future-conditioned behavior learning techniques proposed in \model{} can be seamlessly integrated with DreamerV2-V3, consistently enhancing their overall performance and convergence rate.

\myparagraph{Visual RL with visual distractions.}
However, for complex visual environments with background or even dynamic distractions, it is still challenging to learn effective behavior policies. 
To tackle this problem, some approaches \citep{zhang2021learning,bharadhwaj2022information,wang2022denoised,deng2022dreamerpro} learn a more robust representation by discarding pixel-reconstruction to avoid struggling with the presence of visual noises. 
DreamerPro \citep{deng2022dreamerpro} uses online clustering to learn prototypes from the recurrent states of the world model, eliminating the need for reconstruction. 
Denoised-MDP \citep{wang2022denoised} categorizes system dynamics into four types based on their controllability and relation to rewards, and optimizes the policy model only with information that is both controllable and relevant to rewards.
It is worth noting that \model{} differs significantly from the aforementioned methods in two key ways. 
First, we explicitly model the state transitions of controllable and noncontrollable dynamics in two distinct branches. 
This modular structure empirically facilitates transfer learning between related but distinct domains.
Second, the decoupled world model offers a more versatile method of learning behavior. By previewing possible future states of noncontrollable patterns, we can make informed decisions at present. This also allows us to choose whether or not to incorporate noncontrollable states into our decision-making process, based on our prior knowledge of the specific domain.


\myparagraph{Action-conditioned video prediction.}
Another branch of deep learning solutions to visual control problems is to learn action-conditioned video prediction models \cite{oh2015action,Finn2016Unsupervised,chiappa2017recurrent,wang2021predrnn,babaeizadeh2021fitvid} and then perform Monte-Carlo importance sampling and optimization algorithms, such as the \textit{cross-entropy methods}, over available behaviors \cite{finn2017deep,ebert2018visual,jung2019goal}.  
Hot topics in video prediction mainly include long-term and high-fidelity future frames generation \citep{srivastava2015unsupervised,shi2015convolutional,vondrick2016generating,bhattacharjee2017temporal,wang2017predrnn,villegas2017learning,wichers2018hierarchical,reda2018sdc,oliu2018folded,liu2018dyan,xu2018structure,jin2020exploring,behrmann2021unsupervised}, dynamics uncertainty modeling \cite{babaeizadeh2017stochastic,denton2018stochastic,villegas2019high,kim2019variational,castrejon2019improved,franceschi2020stochastic,wu2021greedy}, object-centric scene decomposition \cite{van2018relational,hsieh2018learning,greff2019multi,zablotskaia2020unsupervised,bei2021learning,greff2017neural,kosiorek2018sequential}, and space-time disentanglement \cite{Villegas2017Decomposing,hsieh2018learning,guen2020disentangling,bodla2021hierarchical}.
The corresponding technical improvements mainly involve the use of more effective neural architectures, novel probabilistic modeling methods, and specific forms of video representations. 
The disentanglement methods are closely related to the world model in \model{}.
They commonly separate visual dynamics into content and motion vectors, or long-term and short-term states.
In contrast, \model{} is designed to learn a decoupled world model based on controllability, which contributes more to the downstream behavior learning process.

\section{Conclusion}

In this paper, we proposed an MBRL framework named \model{}, which mainly tackles the difficulty of vision-based prediction and control in the presence of complex visual dynamics.
Our approach has four novel contributions to world model representation learning and corresponding MBRL algorithms.
First, it learns to decouple controllable and noncontrollable latent state transitions via modular network structures and inverse dynamics. 
Second, it introduces the min-max variance constraints to prevent ``training collapse'', where a single state transition branch captures all information.
Third, it makes long-horizon decisions by rolling out the noncontrollable dynamics into the future and learning their influences on current behavior. 
Fourth, it models the sparse dependency of future noncontrollable dynamics on current controllable dynamics to deal with some practical dynamic environments.
\model{} achieves competitive results on the CARLA autonomous driving task, where other vehicles can be naturally viewed as noncontrollable components, indicating that with the help of decoupled latent states, the agent can make more forward-looking decisions by previewing possible future states in the action-free network branch.
Besides, Our approach was shown to effectively improve the visual control task in a modified DeepMind Control Suite, achieving significant advantages over existing methods in standard, noisy, and transfer learning setups.
%


%



\ifCLASSOPTIONcompsoc
  \section*{Acknowledgments}
\else
  \section*{Acknowledgment}
\fi

This work was supported by the National Natural Science Foundation of China (Grant No. 62250062, 62106144, U19B2035), the Shanghai Municipal Science and Technology Major Project (Grant No. 2021SHZDZX0102), the Fundamental Research Funds for the Central Universities, and the Shanghai Sailing Program (Grant No. 21Z510202133).


\ifCLASSOPTIONcaptionsoff
  \newpage
\fi

\vspace{-30pt}
\begin{IEEEbiography}[{\includegraphics[width=1in,height=1.25in,clip,keepaspectratio]{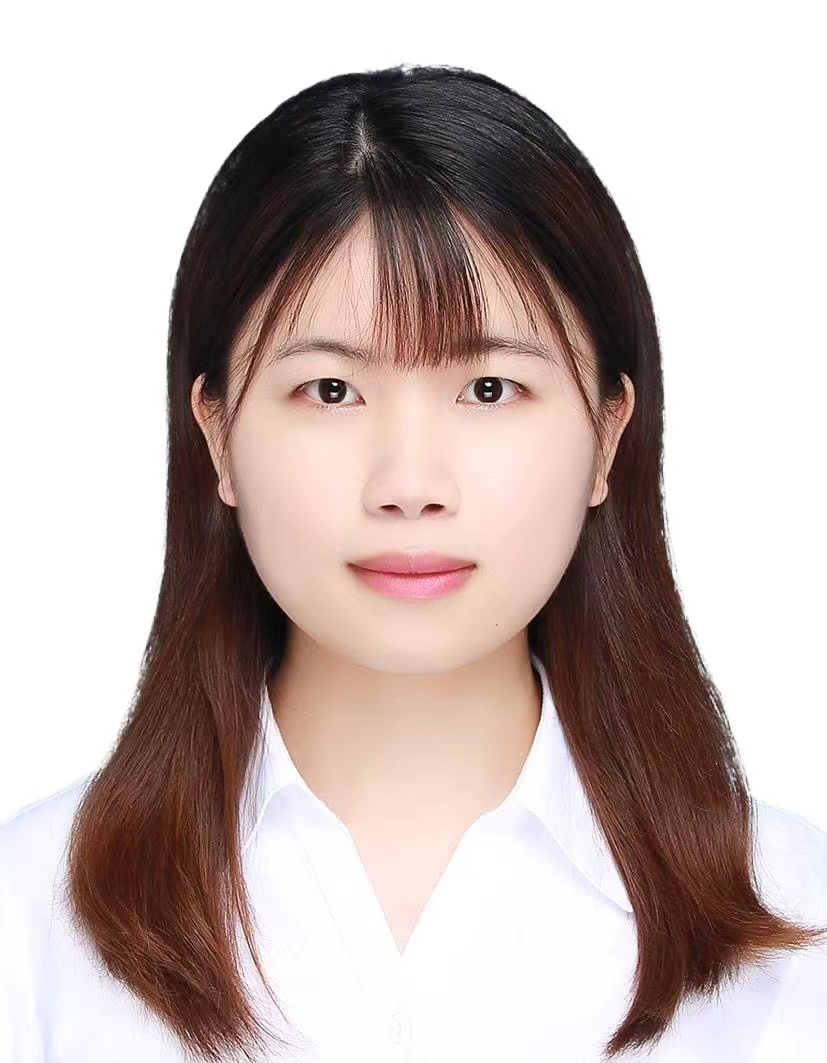}}]{Minting Pan} received the B.E. degree from Hunan University in 2018. She is currently pursuing her PhD degree in Shanghai Jiao Tong University. Her research interests lie on the model-based decision reinforcement learning, especially visual control tasks.
\end{IEEEbiography}

\vspace{-30pt}
\begin{IEEEbiography}[{\includegraphics[width=1in,height=1.25in,clip,keepaspectratio]{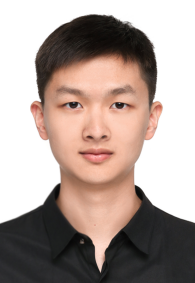}}]{Xiangming Zhu} received the B.E. degree from Shanghai Jiao Tong University in 2022. He is currently pursuing his Master degree in Shanghai Jiao Tong University. His research interests lie on the intersection of machine learning and computer vision, especially vision-based intuitive physics and reinforcement learning.
\end{IEEEbiography}

\vspace{-30pt}
\begin{IEEEbiography}[{\includegraphics[width=1in,height=1.25in,clip,keepaspectratio]{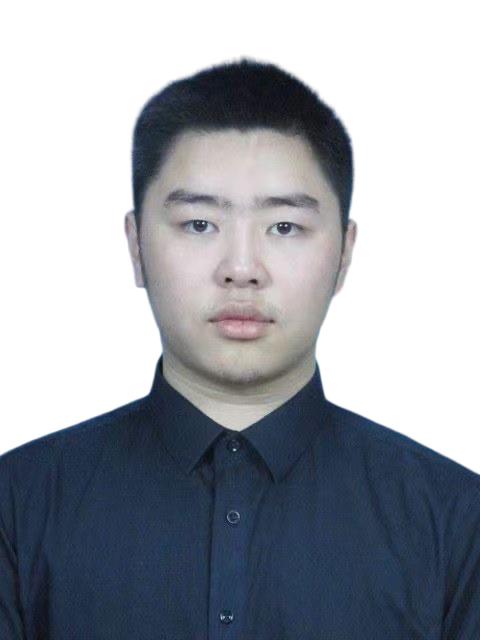}}]{Yitao Zheng} received the B.E. degree from Xidian University in 2023. He is currently purchasing his Ph.D degree in Shanghai Jiao Tong University. His research interest lies in the intersection of computer vision and deep learning, especially model-based visual reinforcement learning.
\end{IEEEbiography}

\vspace{-30pt}
\begin{IEEEbiography}[{\includegraphics[width=1in,height=1.25in,clip,keepaspectratio]{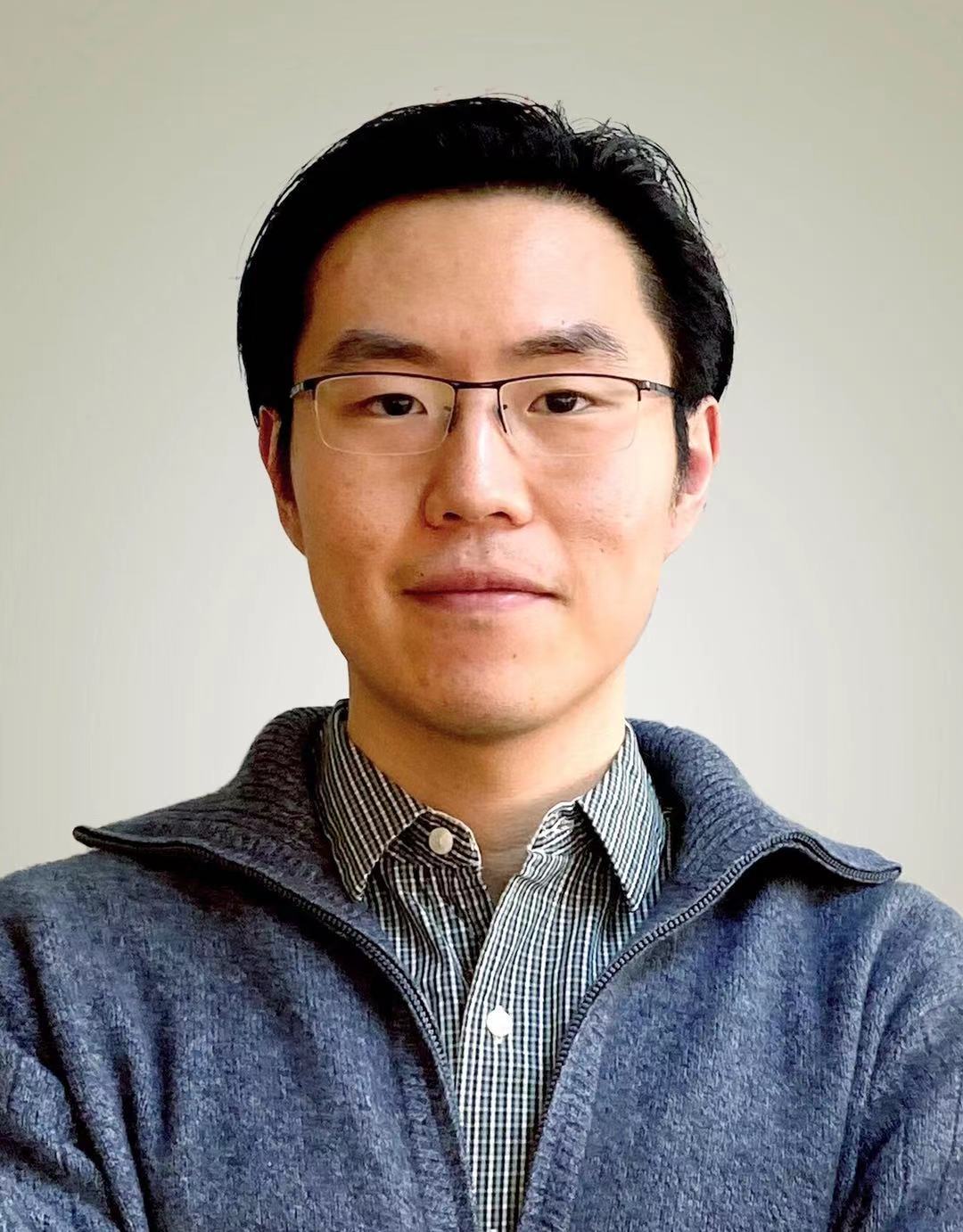}}]{Yunbo Wang} received the B.E. degree from Xi'an Jiaotong University in 2012, and the M.E. and Ph.D. degrees from Tsinghua University in 2015 and 2020. He received the CCF Outstanding Doctoral Dissertation Award in 2020, advised by Philip S. Yu and Mingsheng Long. He is now an assistant professor at the AI Institute and the Department of Computer Science at Shanghai Jiao Tong University. He does research in deep learning, especially predictive learning, spatiotemporal modeling, and model-based decision making. 
\end{IEEEbiography}


\vspace{-30pt}
\begin{IEEEbiography}[{\includegraphics[width=1in,height=1.25in,clip,keepaspectratio]{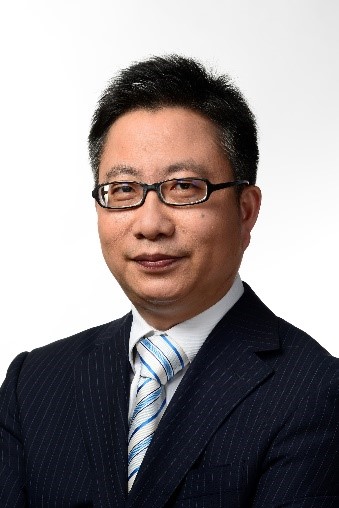}}]{Xiaokang Yang} received the B.S. degree from Xiamen University in 1994, the M.S. degree from the Chinese Academy of Sciences in 1997, and the Ph.D. degree from Shanghai Jiao Tong University in 2000. He is
currently a Distinguished Professor, Shanghai Jiao Tong University, Shanghai, China. His current research interests include visual signal processing and communication, media analysis and retrieval, and pattern recognition. He serves as an Associate Editor of IEEE Transactions on Multimedia.
\end{IEEEbiography}




\end{document}